\documentclass[sigconf,nonacm]{acmart}

\AtBeginDocument{%
  \providecommand\BibTeX{{%
    \normalfont B\kern-0.5em{\scshape i\kern-0.25em b}\kern-0.8em\TeX}}}

\renewcommand\footnotetextcopyrightpermission[1]{} 

\usepackage{mathtools}
\usepackage{amsthm}
\usepackage{bm}
\usepackage{multirow}
\usepackage{algorithm}
\usepackage{algorithmic}
\usepackage{hyperref}
\usepackage{balance}
\usepackage{graphicx}
\usepackage{enumitem}
\usepackage{subfigure}
\DeclareMathOperator*{\argmin}{arg\,min}
\newcommand{\TSP}{\textsc{TSP}}
\newcommand{\CVRP}{\textsc{CVRP}}

\newcommand{\DyNACO}{\textsc{DyNACO}}
\newcommand{\MMAS}{\textsc{MMAS}}

\copyrightyear{2026}
\acmYear{2026}
\setcopyright{cc}
\setcctype{by}
\acmConference[KDD '26]{Proceedings of the 32nd ACM SIGKDD Conference on Knowledge Discovery and Data Mining V.2}{August 09--13, 2026}{Jeju Island, Republic of Korea}
\acmBooktitle{Proceedings of the 32nd ACM SIGKDD Conference on Knowledge Discovery and Data Mining V.2 (KDD '26), August 09--13, 2026, Jeju Island, Republic of Korea}
\acmDOI{10.1145/3770855.3817893}
\acmISBN{979-8-4007-2259-2/2026/08}

\begin{document}

\title{Beyond Static Priors: Dynamic Neural Guidance for \\Large-Scale Ant Colony Optimization}

\author{Dat Thanh Tran}
\orcid{}
\affiliation{%
  \department{Center for AI Research}
  \institution{VinUniversity}
  \city{Hanoi}
  \country{Vietnam}
}
\email{dat.tt3@vinuni.edu.vn}

\author{Van Khu Vu}
\affiliation{%
  \department{College of Engineering and Computer Science}
  \institution{VinUniversity}
  \city{Hanoi}
  \country{Vietnam}
}
\email{khu.vv@vinuni.edu.vn}

\author{Yining Ma}
\authornote{Yining Ma is the corresponding author.}
\affiliation{%
  \department{Laboratory for Information and Decision Systems}
  \institution{Massachusetts Institute of Technology}
  \city{Cambridge, MA}
  \country{USA}
}
\email{yiningma@mit.edu}

\settopmatter{printacmref=true}
\begin{abstract}
Neural-guided Ant Colony Optimization (ACO) suffers from a fundamental training-inference misalignment: policies are typically trained to generate static priors (e.g., heatmaps), yet deployed to guide iterative, long-horizon search processes. In this paper, we present DyNACO, a novel framework that achieves dynamic neural guidance by periodically observing the pheromone distribution and the incumbent solution. To make DyNACO tractable at scale, we pair the policy with a perturbation-based ACO backend and a scope-restricted refinement mechanism that jointly ensure efficacy and stable credit assignment. On TSP, DyNACO scales to 100,000-node instances and outperforms neural baselines while often reducing total runtime compared to the unguided solver. We extend DyNACO to CVRP via a capacity-aware backend, consistently improving the unguided baseline with less than 1\% neural overhead. We further provide in-depth analysis validating the model's generalization capabilities and elucidating why dynamic guidance outperforms static priors. Our work underscores the necessity of aligning neural training with iterative search dynamics in learning-guided optimization. The code is available at \url{https://github.com/shoraaa/DyNACO}.
\end{abstract}

\ccsdesc[500]{Computing methodologies~Sequential decision making}
\ccsdesc[500]{Mathematics of computing~Discrete optimization}

\keywords{Learning-guided Optimization; Neural Combinatorial Optimization; Ant Colony Optimization}

\maketitle

\section{INTRODUCTION}

Combinatorial optimization problems (COPs) such as the Traveling Salesman Problem (\TSP{}) and the capacitated Vehicle Routing Problem (\CVRP{}) have broad applications in logistics and network design~\cite{TSPAPP}.
As NP-hard problems, they are typically tackled with heuristics.
Classical solvers such as LKH-3~\cite{lkhtsp, lkhcvrp} and HGS~\cite{hgs} achieve near-optimal precision through decades of engineering, while recent Machine Learning (ML) based neural combinatorial optimization (NCO)~\cite{MLCO, RL4CO,neuopt,2optlearning} has emerged as a data-driven alternative. However, end-to-end NCO models often struggle with scalability, generalization, and a lack of convergence guarantees~\cite{RL4CO}. Learning-guided optimization (LGO)~\cite{learn2branch, learn2delegate, neurolkh} bridges this gap by integrating neural signals into classical human solvers, which often achieves state-of-the-art performance by effectively combining the strengths of both paradigms~\cite{learn2delegate,deepaco,learn2seg,huang2024contrastive}.

Among the LGO methods, learning to guide Ant Colony Optimization (ACO)~\cite{ACO, maxmin} serves as a foundational approach for steering meta-heuristics~\cite{deepaco,gtgaco}. ACO is a population-based metaheuristic where artificial ants construct solutions via probabilistic transitions biased by pheromone updates toward promising regions. In general, ACO factorizes the transition rule by combining dynamic pheromone updates with heuristic priors. While recent methods successfully learn to initialize such heuristic priors~\cite{deepaco,gfacs} or both heuristics and pheromone matrices~\cite{gtgaco}, they all share a critical flaw: they train a model to produce guidance from instance geometry \emph{only once}, yielding static heatmaps, and then deploy this fixed signal inside a full ACO loop where pheromone dynamics evolve over hundreds of iterations.
The model never observes pheromone during training and has no mechanism to adapt to it at inference.
This \emph{training-inference misalignment} renders the neural signal blind to search progress: it applies identical guidance regardless of whether pheromones are near-uniform (early search) or concentrated around a local optimum (late search).

Motivated by this, we propose \textbf{\DyNACO{}}, a framework that shifts from \emph{static} to \emph{dynamic} neural guidance. While prior methods train the model on a single snapshot of instance geometry and freeze the output, \DyNACO{} trains its policy on the \emph{full search trajectory}: the model repeatedly observes the evolving pheromone distribution and incumbent solution, and learns to emit guidance that is useful at every stage of the search, not just at initialization. To enable this, we design a specialized module to extract representations that summarize the current optimization status. 
We formalize this as a \emph{semi-Markov Decision Process} in which a meta-policy periodically observes search state and emits updated edge-level guidance, and we train \DyNACO{}  with a reinforcement learning algorithm.

Another key limitation of existing neural-guided ACO methods is scalability. To make \DyNACO{} tractable at scale, we pair the policy with a perturbation-based ACO backend on a sparse $K$-nearest-neighbor graph and a \emph{scope-restricted refinement} (SRR) mechanism. Because the resulting perturbations are highly localized, our design delivers three simultaneous benefits: (1) \emph{Stable Credit Assignment}: confining refinement to the neighborhood of the policy's actions preserves the causal link between those actions and post-refinement rewards; (2) \emph{Within-Scope Convergence}: SRR exhaustively applies 2-opt moves around the perturbed edges until no further improvement, ensuring local optimality that guarantees performance; and (3) \emph{Size-Independent Scalability}: the per-ant refinement cost is reduced from $\mathcal{O}(N^2)$ to $\mathcal{O}(M \cdot K)$, making stable training feasible at the 100,000-node scale.

Extensive experiments on TSP and CVRP demonstrate that \DyNACO{} significantly outperforms existing neural ACO baselines under matched budgets, and surpasses established classical and neural baselines, especially on real-world large-scale instances. On TSP, neural guidance actually \emph{reduces} total runtime by 23-33\% over the unguided solver, as better-targeted perturbations accelerate local-search convergence; on CVRP, neural overhead stays below 1\%, and end-to-end wall-clock overhead remains within 1-3\% at large scales, with consistent quality gains. Finally, we conduct in-depth analyses revealing the learned guidance patterns and explaining their superiority over static priors: in particular, the policy actively counteracts ACO stagnation~\cite{stagnation} by suppressing over-reinforced edges and redirecting search toward under-explored regions. Our work underscores the necessity of aligning neural training with iterative search dynamics in learning-guided optimization.

Our contributions are as follows:
(i)~we propose \DyNACO{}, the first framework for dynamic neural guidance in ACO, formulated as a semi-MDP that lets a policy inject updated guidance at regular intervals rather than committing to a single static prediction;
(ii)~we introduce SRR, confining local search to the perturbation neighborhood to preserve credit assignment and achieve 2-opt optimality with efficiency;
(iii)~we show that \DyNACO{} scales to 100K nodes while reducing total runtime on TSP and adding $<$1\% overhead on CVRP, with zero-shot transfer on TSPLIB and CVRPlib confirming cross-scale and cross-distribution generalization.

\section{RELATED WORK}
\label{sec:related-nco}

\smallskip\noindent\textbf{End-to-end neural solvers.} Typical neural solvers include auto-regressive neural models~\cite{attentionmodel,pomo} and non-auto-regressive neural models~\cite{bq,lehd,invit,sigd,l2c}, which learn constructive policies directly from data, automating solution generation and enabling batch optimization. In principle, NCO can optimize with minimal manual re-engineering, since the learned policy adapts to new instances without re-designing heuristics. In practice, however, they suffer from optimality gaps on hard instances, difficulty generalizing across problem sizes and distributions, and inefficiency on large-scale instances. For example, scaling beyond a few thousand nodes remains challenging due to quadratic attention complexity and error accumulation along long construction trajectories (Appendix~\ref{app:related}).

\smallskip\noindent\textbf{Learning-guided optimization.}
These methods~\cite{learn2branch, learn2delegate, neurolkh} combine the strengths of both paradigms: the convergence guarantees and problem-specific structure of classical solvers, together with the data-driven adaptability of neural methods.
Rather than replacing the solver, a neural network supplies signals while the solver retains its algorithmic backbone, mitigating the scalability limitations of end-to-end NCO and the rigidity of hand-crafted rules. 

\smallskip\noindent\textbf{Neural Guided ACO.}  Ant Colony Optimization (ACO)~\cite{ACO, maxmin} is a population-based metaheuristic that explores solution space through pheromone-guided sampling. Recent works integrate neural components into ACO through diverse mechanisms: DeepACO \cite{deepaco} and GTG-ACO \cite{gtgaco} employ GNNs and Transformers to learn initialization of the heuristic matrices and/or the pheromone matrices, GFACS \cite{gfacs} leverages GFlowNets for constructive sampling, and HeatACO \cite{heataco} utilizes pre-computed heatmaps as probabilistic priors. Despite these architectural distinctions, existing methods suffer from two primary limitations: (1) they rely on static guidance that remains invariant to the search trajectory, and (2) their scalability to large-scale instances is still restricted.


\section{METHODOLOGY}
\label{sec:method}

\begin{figure*}[t]
\centering
\includegraphics[width=\textwidth]{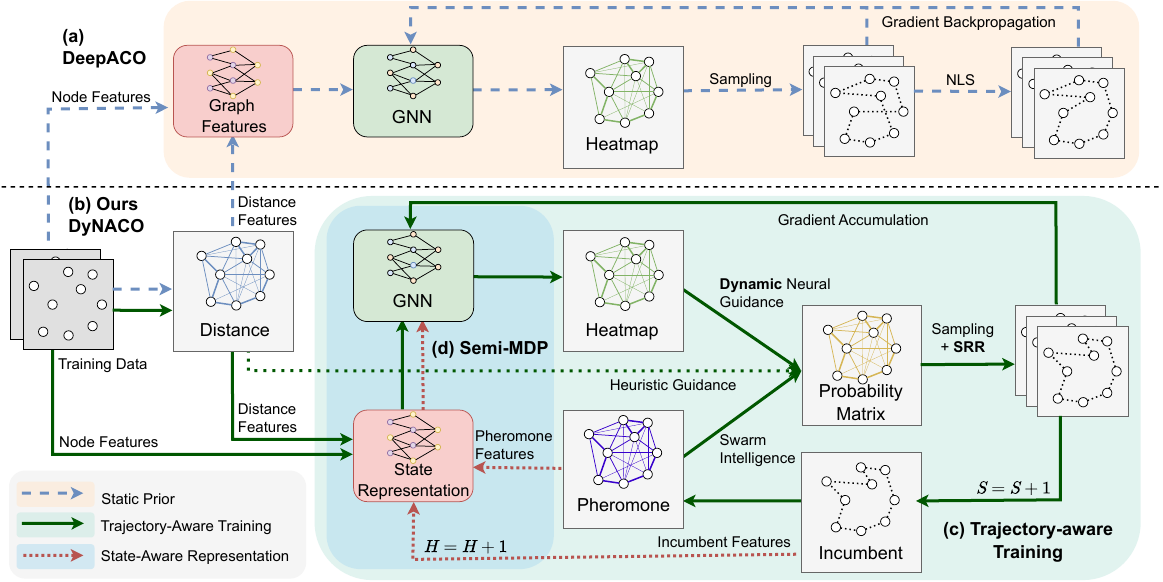}
\caption{\emph{Comparison of training paradigms and our \DyNACO{}.} (a) The original DeepACO paradigm~\cite{deepaco, gfacs, gtgaco}. (b) Our \DyNACO{} framework for dynamic guidance. (c) Trajectory-aware training that aligns with iterative search dynamics. (d) State-aware representation conditioning the policy on the evolving search state.}
\label{fig:guidance-loop}
\end{figure*}

Standard ACO constructs solutions using transition probabilities $p(j \mid i) \propto [\tau_{ij}]^\alpha [\eta_{ij}]^\beta$, relying on an evolving pheromone matrix $\tau_{ij}$ but a strictly static heuristic visibility $\eta_{ij}$. This section presents \DyNACO{}, injecting \emph{dynamic neural guidance} into ACO (Figure~\ref{fig:guidance-loop}) based on a state-aware representation learning trained by a trajectory-aware objective that optimizes expected cost across the full search history, aligning the training signal with iterative search dynamics. Moreover, \DyNACO{} features a perturbation-based ACO backend and a scope-restricted refinement mechanism for scalability. To evaluate \DyNACO{}, we apply it to the Euclidean \TSP{} on $N$ nodes, seeking a tour $\pi$ that minimizes total cost $C(\pi)$, as well as its capacity-constrained variant, \CVRP{}, where we design a customized capacity-aware perturbation backend for feasibility handling. Regarding node features, we adapt those static features from DeepACO~\cite{deepaco}, and add a set of dynamic edge features derived from the evolving pheromone statistics and the incumbent solution. 

\subsection{Neural ACO Guidance as a Semi-MDP}
\label{sec:mdp}
Prior neural-guided ACO methods~\cite{deepaco,gfacs,gtgaco} treat guidance as a \emph{one-shot prediction}: a network observes the instance graph and emits a static heatmap, trained from a single pheromone-free sample of ants, after which the solver runs a full ACO loop with pheromone dynamics to completion without further interaction. The model is therefore trained without pheromone dynamics. We address this by formalizing \DyNACO{} as a two-level hierarchy in which a neural \emph{meta-policy} $\pi_\theta$ \emph{repeatedly} observes the macro-state and emits updated guidance, while the ACO backend executes $S$ iterations between observations. Such a semi-MDP formulation enables:

\begin{enumerate}
\item \emph{Amortized policy updates.} The neural policy outputs actions only once every $S$ iterations and remains fixed during this interval. This amortizes overhead and effectively reduces variance during policy-gradient training.
\item \emph{Efficient PPO replay.} By framing the guidance process as a semi-MDP, we avoid tracking low-level internal ACO ant steps. To compute PPO importance ratios for off-policy reinforcement learning updates, we only need to store the macro-states at the update steps: the pheromone snapshots, incumbent features, and action traces.
\item \emph{Trajectory-aware objective.} Instead of a myopic single-step reward as used in the literature, we minimize the long-term cumulative cost $\sum_{h=1}^H C_h$ across the full MDP trajectory.
\end{enumerate}

Formally, we model this as a \emph{semi-MDP with fixed-duration macro-actions} as follows:
\begin{itemize}[topsep=0pt]
\item \emph{State} $\mathcal{S}_h = (G, \tau^{(h,0)}, b^{(h)})$: At update step $h$, the macro-state comprises the static $K$-nearest-neighbor candidate graph $G$, the pheromone matrix $\tau^{(h,0)} \in \mathbb{R}^{N \times K}$ at the start of the interval (indexed as $h,0$), and the current solution $b^{(h)}$.
\item \emph{Action} $a_h = z_\theta(\mathcal{S}_h) \in \mathbb{R}^{N \times K}$: A residual logit matrix emitted by the policy that shapes the pheromone matrix.
\item \emph{Transition} $P(\mathcal{S}_{h+1} \mid \mathcal{S}_h, a_h)$: The environment executes $S$ internal iterations of ACO (comprising ant sampling, scope-restricted refinement, and pheromone updates), functioning as a stochastic, state-dependent transition kernel.
\item \emph{Cost} $C_h = \frac{1}{S \cdot m} \sum_{s=1}^{S} \sum_{a=1}^{m} C(\hat{\pi}_{a,s})$: The average post-refinement tour cost across all $m$ ants over the $S$ iterations. Our objective is to minimize the expected cumulative cost $\sum_{h=1}^H C_h$.
\end{itemize}

\subsection{The Perturbative Environment}
\label{sec:backend}

Before defining the neural policy, we introduce the environment it operates within: a perturbation-based ACO backend designed to generate stochastic transitions between macro-states. This environment relies on two core mechanisms to make policy-gradient training viable at scale: \emph{perturbative sampling} to break the $O(N)$ decision horizon, and \emph{Scope-Restricted Refinement (SRR)} to guarantee stable credit assignment without sacrificing local optimality.

\smallskip\noindent\textbf{Perturbative Sampling.}
In prior neural-guided ACO frameworks~\cite{deepaco,gfacs,gtgaco}, each ant constructs a tour from scratch. This entails a sequence of $N$ sequential edge selections, which becomes prohibitively expensive at large scales and suffers from compounding errors. Furthermore, this full-tour construction provides no natural anchor for localized refinement. To overcome this, we adopt a perturbation-based ACO operating on a sparse $K$-nearest-neighbor candidate graph. Instead of building from scratch, each ant initializes from the current incumbent solution and modifies only a small subset of edges. For \TSP{}, we adapt the backend from~\cite{faco}; for \CVRP{}, we design a novel capacity-aware variant (detailed in Appendix~\ref{app:backend}).

At sampling iteration $s$ of guidance update $h$, transition probabilities combine pheromones, heuristics, and neural guidance via a \emph{guidance weight} $\gamma$:
\begin{equation}
\log p^\theta(j \mid i; h, s) = \alpha \log \tau^{(h,s)}_{ij} + \beta \log \eta_{ij} + \gamma\, z_{ij}(\mathcal{S}_h).
\label{eq:held-logits}
\end{equation}
During training $\gamma{=}1$ (constant forcing); at inference $\gamma$ may be annealed (see Section \ref{sec:inference}). The policy outputs $z_\theta(\mathcal{S}_h)$ once per guidance update and holds it constant for $S$ sampling iterations, decoupling neural guidance (updated at guidance-update boundaries) from pheromone dynamics (updated at every sampling iteration). Crucially, prior neural-ACO methods~\cite{deepaco,gfacs,gtgaco} replace the heuristic term~$\eta_{ij}$ with neural output, discarding hand-crafted domain knowledge. We instead retain both $\tau_{ij}$ and $\eta_{ij}$, and introduce neural logits~$z_{ij}$ as a third, additive term in Eq.~\eqref{eq:held-logits}. This preserves the domain knowledge encoded in $\eta$ while letting the network provide complementary, state-dependent guidance.

Formally, each ant initializes from the incumbent solution $b^{(h)}$ and executes up to $M$ relocation local search steps that introduce new edges. At each step, the ant samples a successor node $v$ from the set of unvisited candidates according to Eq.~\eqref{eq:held-logits}; edges $(u,v)$ absent from the incumbent are recorded as anchors for the subsequent scope-restricted refinement (SRR) mechanism. The trajectory log-probability decomposes over stochastic decisions:
\begin{equation}
\log \pi_\theta(\pi_{a,s} \mid \mathcal{S}_h) = \sum_{t=1}^{T_a} \log p^\theta(v_t \mid u_t; h, s),
\label{eq:logp-trajectory}
\end{equation}
where $T_a \le M$ denotes the number of stochastic edge selections, i.e., steps at which multiple candidates were feasible. This formulation reflects a fundamental shift in problem framing: constructive ACO selects all $N$ edges comprising a tour, whereas perturbative ACO identifies the subset of $M \ll N$ edges to modify. Consequently, the decision horizon is bounded by $M$ rather than $N$, preventing the error accumulation inherent in $\mathcal{O}(N)$-length sequential construction and stabilizing policy learning at scale. The resulting localized perturbations further enable the SRR procedure described below.

\smallskip\noindent\textbf{Scope-Restricted Refinement.}
\label{sec:localized-ls}
In neural-guided ACO~\cite{deepaco,gfacs,gtgaco}, local search is part of the training environment: the post-refinement cost is the reward signal used for policy-gradient updates. This creates a credit-assignment trade-off. If local search is too weak, sampled tours remain under-refined and the reward is noisy. If local search is too strong and unconstrained, it can rewrite most of the policy's perturbation, so the final cost reflects the local-search heuristic rather than the neural decisions. We refer to this failure mode as \emph{gradient washout}.

\begin{figure}[t]
    \centering
    \includegraphics[width=\columnwidth]{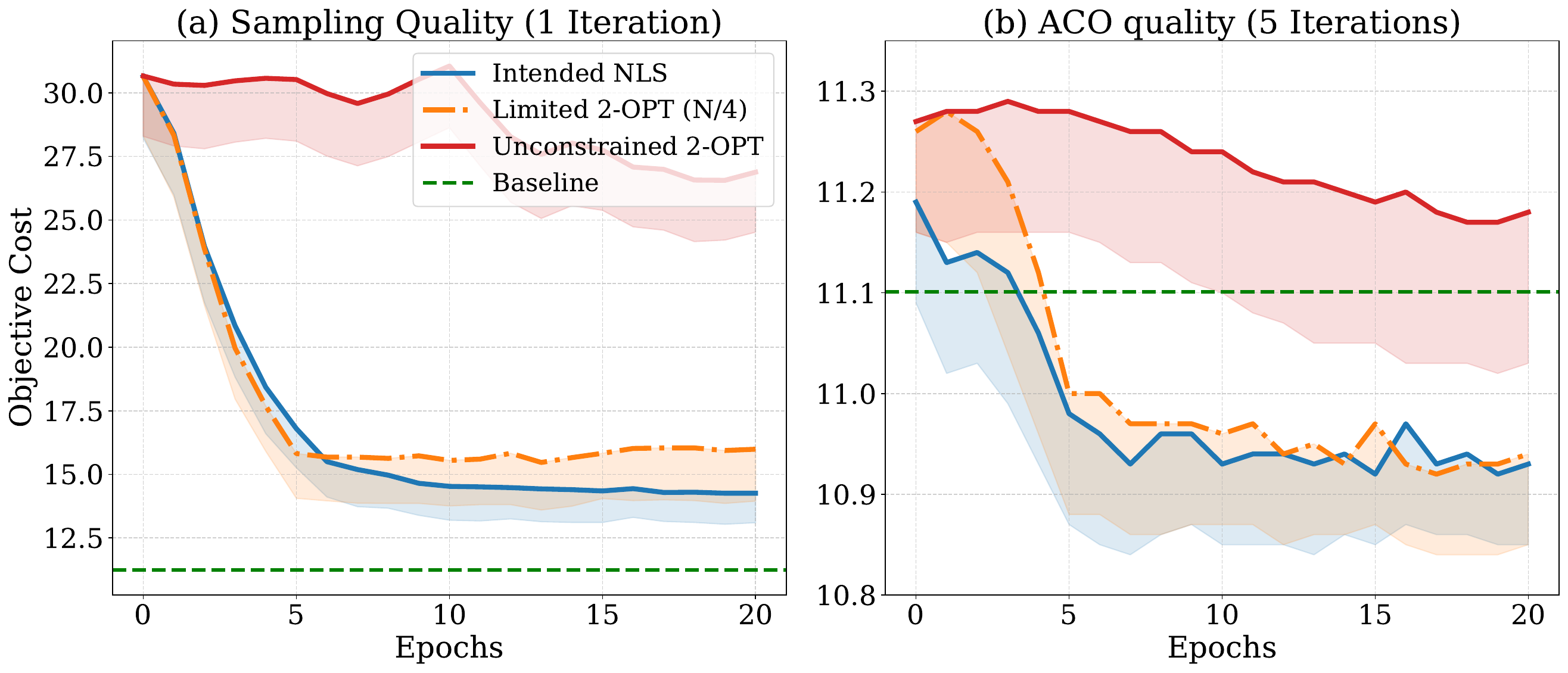}
    \caption{Credit-assignment trade-off under different local-search strategies when training DeepACO on TSP-200.}
    \label{fig:credit_assignment}
\end{figure}

Figure~\ref{fig:credit_assignment} illustrates this trade-off using DeepACO on TSP-200 under three refinement strategies. With \emph{Full 2-opt} (red), the policy fails to converge because the local search heavily rewrites the sampled tour and severs the causal link between actions and rewards. \emph{Limited 2-opt} (orange) preserves more of the learning signal by stopping after $N/4$ swaps, but this early stopping bottlenecks solution quality. \emph{Neural Local Search}~\cite{deepaco} (blue) improves quality by using the neural policy to score candidate swaps, but it is not scalable because it requires neural evaluations inside the refinement loop.

Moreover, we note that scalability imposes an additional constraint. Standard 2-opt costs $\mathcal{O}(N^2)$ per tour, which is prohibitive when refinement must be applied to every ant at every iteration. Classical large-scale ACO methods~\citep{partialACO, P-ACO} reduce this cost by applying local search only occasionally or only to elite ants. This is unsuitable for policy-gradient training, where per-ant advantages require post-refinement costs for all sampled ants. Thus, the refinement operator must satisfy three requirements simultaneously: it must preserve the causal link between policy actions and rewards, refine tours to convergence rather than stop at an arbitrary budget, and have cost independent of the full instance size.

We satisfy these requirements using Scope-Restricted Refinement (SRR) inspired by ~\citet{faco}. The key observation is that perturbation-based ants start from the incumbent solution, which is already locally optimized under the candidate graph. An ant modifies only a small set of edges, so any newly introduced sub-optimality is localized around those perturbations. SRR initializes a checklist with the endpoints of the perturbed edges and repeatedly applies improving candidate-graph 2-opt moves involving checked nodes, reactivating affected neighbors after each accepted move. In this sense, SRR behaves like a localized repair process that propagates outward from the perturbation region until no improving move remains within its scope.

Specifically, SRR provides the three properties required for training. First, because refinement is restricted to the neighborhood of the policy's perturbations, the post-refinement cost remains causally tied to the sampled actions, preserving credit assignment. Second, unlike truncated local search, SRR runs to convergence within its scope; assuming the incumbent was already candidate-graph 2-opt optimal, this restores candidate-graph 2-opt optimality after the perturbation. Third, the refinement cost scales with the perturbation size, $\mathcal{O}(M \cdot K)$, rather than the instance size, $\mathcal{O}(N^2)$, allowing SRR to be applied to every ant at every iteration even at $N{=}100$K. A formal optimality argument is given in Appendix~\ref{app:credit}.

\smallskip\noindent\textbf{Stabilized Pheromone Dynamics.}
\label{sec:smoothed-mmas}
Lastly, we note that \MMAS{}~\cite{maxmin} bounds pheromones to $[\tau_{\min}, \tau_{\max}]$ to prevent premature convergence, but couples these bounds to solution cost, creating non-stationary input distributions. We instead adopt fixed bounds ($\tau_{\max}{=}1$, $\tau_{\min}{=}1/K$) and update via convex interpolation:
\begin{equation}
\tau^{(h,s+1)}_{ij} \leftarrow (1-\rho)\tau^{(h,s)}_{ij} + \rho \cdot \mathcal{T}_{ij},
\label{eq:smoothed-mmas}
\end{equation}
where $\mathcal{T}_{ij} = \tau_{\max}$ if $(i,j)$ is in the best solution, else $\tau_{\min}$.
This yields scale-invariant state representations and stabilizes training.
While this simplification may reduce the solution quality of the \emph{unguided} solver compared to highly tuned \MMAS{} dynamics, our ablations (Section \ref{sec:ablations}) show that DyNACO with stabilized bounds outperforms both (i)~unguided ACO with standard \MMAS{} updates and (ii)~neural-guided ACO retaining the original update rule.

\subsection{Dynamic Neural Guidance}
\label{sec:agent}

\smallskip\noindent\textbf{State-Aware Representation.}
\label{sec:state}
Prior neural-ACO methods condition solely on \emph{static} instance geometry, yielding guidance that cannot adapt to the solver's evolving state. The semi-MDP state $\mathcal{S}_h = (G,\, \tau^{(h,0)},\, b^{(h)})$ defined in Section \ref{sec:mdp} isolates the two quantities that evolve between guidance updates: the pheromone field~$\tau$ and the incumbent solution~$b$. These are precisely the signals inherent to the ACO process; conditioning on them enables the semi-MDP formulation and renders the guidance \emph{dynamic}. We retain the static node features used by DeepACO~\cite{deepaco} for each problem and introduce dynamic edge features derived from $\tau$ and $b$, capturing local pheromone concentration, convergence status, and incumbent topology (Appendix~\ref{app:state-features}). These features are processed by a 12-layer GNN encoder~\cite{deepaco} with interleaved node and edge message-passing. All inputs are bounded by $\mathcal{O}(N \cdot K)$, ensuring that state-aware conditioning remains highly tractable at scale.

The policy outputs a scalar guidance score $z_{ij}$ for each candidate edge via a 3-layer MLP decoder.
The shaped transition kernel adds these logits in log-space (Eq.~\ref{eq:held-logits}), corresponding to a KL-regularized policy where $p^\theta(\cdot\mid i) \propto q(j\mid i)\exp(z_{ij})$.
Because $\tau_{\min} > 0$ and logits remain finite, every candidate retains positive probability; the policy redistributes mass but cannot exclude edges.
\begin{algorithm}[t]
\caption{PPO training for \DyNACO{}.}
\label{alg:train}
\begin{algorithmic}[1]
\REQUIRE $\theta$ (policy params), $E$ (epochs), $H$ (outer steps), $S$ (inner steps), $K_{\text{PPO}}$ (PPO epochs), $m$ (ants)
\FOR{$e = 1 \to E$}
    \STATE $x \sim \mathcal{X}$; \, $\tau^{(1,0)} \gets \tau_{\max}$; \, $b^{(1)} \gets \textsc{Greedy}(x)$
    \FOR{$h = 1 \to H$}
        \STATE $z \gets z_{\theta_{\text{old}}}(\mathcal{S}_h)$; \, $\mathcal{D}_h \gets \emptyset$
            \COMMENT{Outer: emit guidance}
        \FOR{$s = 1 \to S$}
            \STATE $\{\pi_{a,s}\}_{a=1}^m \gets \textsc{Sample}(p^\theta, \tau^{(h,s)}, z, b^{(h)})$
                \COMMENT{Inner: ACO sampling}
            \STATE $\mathcal{D}_h \gets \mathcal{D}_h \cup \{(\tau^{(h,s)}, \{\pi_{a,s}\}, \{C(\hat{\pi}_{a,s})\})\}$
            \STATE $a^* \gets \argmin_a C(\hat{\pi}_{a,s})$; \, $b^{(h)} \gets \min(b^{(h)}, \hat{\pi}_{a^*,s})$
            \STATE $\tau^{(h,s+1)} \gets (1{-}\rho)\tau^{(h,s)} + \rho \cdot \Delta\tau(b^{(h)})$
        \ENDFOR
        \STATE $\tau^{(h+1,0)} \gets \tau^{(h,S)}$; \, $b^{(h+1)} \gets b^{(h)}$
        \FOR{$k = 1 \to K_{\text{PPO}}$}
            \STATE $\theta \gets \theta - \nabla_\theta \mathcal{L}^{\text{PPO}}(\theta; \mathcal{D}_h, z_\theta(\mathcal{S}_h))$
                \COMMENT{PPO update}
        \ENDFOR
    \ENDFOR
\ENDFOR
\end{algorithmic}
\end{algorithm}

\smallskip\noindent\textbf{Trajectory-Aware Training.}
\label{sec:training}
Prior methods~\cite{deepaco,gfacs,gtgaco} optimize a \emph{single-step} objective: $\min_\theta \mathbb{E}_{\pi \sim p_\theta(\cdot \mid G)} [ C(\pi) ]$, where ants sample once from the neural heuristic alone.
At inference, however, the same static heatmap is inserted into a full ACO loop where pheromone fields evolve over hundreds of iterations.
This discrepancy precludes learning adaptive behavior: because training never exposes the model to pheromone dynamics, the heatmap inevitably conflicts with the pheromone landscape in later iterations.

We instead optimize the expected per-iteration post-SRR cost across the full trajectory.
Let $\hat{\pi}_{a,s} = \mathrm{LS}(\pi_{a,s})$ denote the tour produced by ant $a$ at sampling iteration $s$ after local search:
\begin{equation}
J(\theta) = \mathbb{E} \left[ \frac{1}{I} \sum_{h=1}^H \sum_{s=1}^S \frac{1}{m} \sum_{a=1}^m C(\hat{\pi}_{a,s}) \right].
\end{equation}
This \emph{trajectory-aware} objective averages cost over all $I = H \times S$ iterations, training the policy to remain effective from early exploration through late exploitation and minimizing the area under the cost curve rather than a single endpoint.

Under perturbative sampling each ant modifies at most $M$ edges, and the log-probability of its trajectory is the sum over stochastic edge choices.
Because the number of stochastic decisions $T_a$ varies across ants (deterministic steps with a single feasible candidate contribute $\log p = 0$), we normalize by decision count: $\bar{\ell}_{a,s} = \frac{1}{T_a} \sum_{t=1}^{T_a} \log p^\theta(v_t \mid u_t)$.
This per-decision normalization prevents ants with more stochastic steps from dominating the gradient.
We compute a per-iteration baseline $b_s = \frac{1}{m}\sum_{a=1}^{m} C(\hat{\pi}_{a,s})$ and define advantage as $A_{a,s} = b_s - C(\hat{\pi}_{a,s})$, so that better-than-average tours receive positive advantage.
For PPO we normalize advantages across the batch: $\hat{A}_{a,s} = (A_{a,s} - \mu_A) / (\sigma_A + \epsilon)$.

\begin{table*}[t]
\centering
\caption{Comparative results on synthetic \TSP{} and \CVRP{} instances. OOM: out of memory.}
\label{tab:main-results}
\small
\renewcommand{\arraystretch}{0.8}
\resizebox{\textwidth}{!}{
\begin{tabular}{lcccccccccc}
\toprule
& \multicolumn{2}{c}{\TSP{}1K} & \multicolumn{2}{c}{\TSP{}5K} & \multicolumn{2}{c}{\TSP{}10K} & \multicolumn{2}{c}{\TSP{}50K} & \multicolumn{2}{c}{\TSP{}100K} \\
\cmidrule(lr){2-3} \cmidrule(lr){4-5} \cmidrule(lr){6-7} \cmidrule(lr){8-9} \cmidrule(lr){10-11}
Method & Obj.\ (Gap) & Time & Obj.\ (Gap) & Time & Obj.\ (Gap) & Time & Obj.\ (Gap) & Time & Obj.\ (Gap) & Time \\
\midrule
LKH3 & 23.12 (0.00\%) & 1.70m & 50.97 (0.00\%) & 12.00m & 71.78 (0.00\%) & 33.00m & 159.93 (0.00\%) & 10.00h & 225.99 (0.00\%) & 25.00h \\
\midrule
H-TSP & 24.66 (6.66\%) & 48.00s & 55.16 (8.22\%) & 1.20m & 77.75 (8.32\%) & 2.20m & OOM & OOM & OOM & OOM \\
GLOP & 23.78 (2.85\%) & 10.20s & 53.15 (4.28\%) & 1.00m & 75.04 (4.54\%) & 1.90m & 168.09 (5.10\%) & 1.50m & 237.61 (5.14\%) & 3.90m \\
INViT-3V greedy & 24.66 (6.66\%) & 9.00s & 54.49 (6.91\%) & 1.20m & 76.85 (7.06\%) & 3.70m & 171.42 (7.18\%) & 1.30h & 242.26 (7.20\%) & 5.00h \\
L2C-Insert greedy & 24.22 (4.75\%) & 0.16s & -- & - & 77.34 (7.75\%) & 3.98s & 171.78 (7.41\%) & 19.80s & 242.76 (7.42\%) & 39.49s \\
SIL greedy & 23.57 (1.94\%) & 0.20s & 52.59 (3.18\%) & 5.20s & 74.69 (4.05\%) & 20.10s & 168.50 (5.36\%) & 7.70m & 239.84 (6.13\%) & 33.00m \\
POMO aug×8 & 32.51 (40.61\%) & 4.10s & 87.72 (72.10\%) & 8.60m & OOM & OOM & OOM & OOM & OOM & OOM \\
BQ bs16 & 23.43 (1.34\%) & 13.00s & 58.27 (14.32\%) & 24.00s & OOM & OOM & OOM & OOM & OOM & OOM \\
SIGD bs16 & 23.36 (1.04\%) & 17.30s & 55.77 (9.42\%) & 30.50m & OOM & OOM & OOM & OOM & OOM & OOM \\
LEHD RRC$_{1000}$ & 23.29 (0.74\%) & 3.30m & 54.43 (6.79\%) & 8.60m & 80.90 (12.71\%) & 18.60m & OOM & OOM & OOM & OOM \\
L2C-Insert (I$_{1000}$) & 23.23 (0.48\%) & 21.75s & -- & - & 73.27 (2.08\%) & 1.04m & 166.06 (3.83\%) & 1.30m & 237.11 (4.92\%) & 1.63m \\
SIL PRC$_{10}$ & 23.40 (1.19\%) & 0.90s & 52.36 (2.73\%) & 5.10s & 73.99 (3.08\%) & 10.00s & 166.69 (4.23\%) & 1.33m & 235.38 (4.16\%) & 3.00m \\
SIL PRC$_{1000}$ & 23.21 (0.38\%) & 1.50m & 51.67 (1.37\%) & 9.40m & 73.08 (1.81\%) & 17.00m & 163.95 (2.51\%) & 1.38h & 231.52 (2.45\%) & 2.60h \\
\midrule
ACO I$_{1000}$ & 23.31 (0.83\%) & 0.54s & 52.12 (2.26\%) & 1.54s & 73.80 (2.82\%) & 2.66s & 168.17 (5.15\%) & 11.86s & 241.26 (6.76\%) & 25.02s \\
ACO I$_{2000}$ & 23.28 (0.70\%) & 1.08s & 51.98 (1.98\%) & 3.11s & 73.54 (2.46\%) & 5.30s & 166.25 (3.95\%) & 23.07s & 237.53 (5.10\%) & 47.56s \\
ACO I$_{5000}$ & 23.25 (0.55\%) & 2.72s & 51.88 (1.78\%) & 7.74s & 73.33 (2.15\%) & 13.41s & 164.90 (3.11\%) & 56.43s & 234.33 (3.69\%) & 1.89m \\
ACO I$_{10000}$ & 23.23 (0.48\%) & 5.44s & 51.81 (1.65\%) & 15.59s & 73.23 (2.02\%) & 26.81s & 164.38 (2.78\%) & 1.87m & 233.04 (3.12\%) & 3.72m \\
\midrule
DyNACO I$_{1000}$ & 23.24 (0.52\%) & 0.37s & 51.63 (1.30\%) & 1.19s & 72.92 (1.59\%) & 1.94s & 167.36 (4.64\%) & 9.50s & 241.20 (6.73\%) & 21.52s \\
DyNACO I$_{2000}$ & 23.21 (0.39\%) & 0.73s & 51.52 (1.08\%) & 2.23s & 72.67 (1.24\%) & 3.72s & 164.58 (2.91\%) & 17.31s & 236.68 (4.73\%) & 38.45s \\
DyNACO I$_{5000}$ & 23.18 (0.27\%) & 1.85s & 51.41 (0.86\%) & 5.30s & 72.48 (0.97\%) & 9.06s & 162.66 (1.71\%) & 40.64s & 232.07 (2.69\%) & 1.48m \\
DyNACO I$_{10000}$ & \textbf{23.17} (\textbf{0.20\%}) & 3.68s & \textbf{51.35} (\textbf{0.75\%}) & 10.42s & \textbf{72.37} (\textbf{0.82\%}) & 17.89s & \textbf{161.93} (\textbf{1.25\%}) & 1.33m & \textbf{230.28} (\textbf{1.90\%}) & 2.86m \\
\midrule
\midrule
& \multicolumn{2}{c}{\CVRP{}1K} & \multicolumn{2}{c}{\CVRP{}5K} & \multicolumn{2}{c}{\CVRP{}10K} & \multicolumn{2}{c}{\CVRP{}50K} & \multicolumn{2}{c}{\CVRP{}100K} \\
\cmidrule(lr){2-3} \cmidrule(lr){4-5} \cmidrule(lr){6-7} \cmidrule(lr){8-9} \cmidrule(lr){10-11}
Method & Obj.\ (Gap) & Time & Obj.\ (Gap) & Time & Obj.\ (Gap) & Time & Obj.\ (Gap) & Time & Obj.\ (Gap) & Time \\
\midrule
HGS & 36.29 (0.00\%) & 2.50m & 89.74 (0.00\%) & 2.00h & 107.40 (0.00\%) & 5.00h & 267.73 (0.00\%) & 8.10h & 476.11 (0.00\%) & 24.00h \\
GLOP-G (LKH3) & 39.50 (8.85\%) & 1.30s & 98.90 (10.21\%) & 6.80s & 116.28 (8.27\%) & 11.20s & OOM & OOM & OOM & OOM \\
\midrule
POMO aug×8 & 84.89 (133.92\%) & 4.80s & 393.27 (338.23\%) & 11.00m & OOM & OOM & OOM & OOM & OOM & OOM \\
LEHD RRC$_{1000}$ & 37.43 (3.14\%) & 3.40m & 101.07 (12.63\%) & 31.00m & 138.73 (29.17\%) & 41.00m & OOM & OOM & OOM & OOM \\
BQ bs16 & 38.17 (5.18\%) & 14.00s & 104.40 (16.34\%) & 2.60m & OOM & OOM & OOM & OOM & OOM & OOM \\
SIGD bs16 & 39.15 (7.88\%) & 17.30s & 103.46 (15.29\%) & 1.91m & 131.48 (22.42\%) & 3.97m & 477.43 (78.33\%) & 25.90m & OOM & OOM \\
INViT-3V greedy & 42.75 (17.80\%) & 11.40s & 109.85 (22.41\%) & 1.40m & 141.41 (31.67\%) & 4.20m & 402.05 (50.17\%) & 2.90h & 688.80 (44.67\%) & 8.30h \\
LEHD greedy & 38.91 (7.22\%) & 0.80s & 105.61 (17.68\%) & 1.56m & 146.24 (36.16\%) & 11.85m & OOM & OOM & OOM & OOM \\
L2C-Insert greedy & 39.69 (9.36\%) & 0.41s & 109.45 (21.96\%) & 3.30s & 147.98 (37.78\%) & 6.64s & 419.51 (56.69\%) & 33.83s & 722.17 (51.68\%) & 1.14m \\
L2C-Insert I$_{1000}$ & 38.33 (5.63\%) & 43.59s & 103.60 (15.45\%) & 1.21m & 135.77 (26.42\%) & 1.27m & 389.86 (45.62\%) & 1.74m & 692.95 (45.54\%) & 2.38m \\
SIL PRC$_{10}$ & 37.93 (4.52\%) & 0.70s & 93.92 (4.66\%) & 3.90s & 112.17 (4.44\%) & 6.80s & 285.20 (6.53\%) & 28.00s & 496.24 (4.23\%) & 59.00s \\
SIL PRC$_{1000}$ & 37.28 (2.73\%) & 1.50m & \textbf{90.81} (\textbf{1.19\%}) & 8.80m & \textbf{106.69} (\textbf{-0.66\%}) & 15.20m & \textbf{262.82} (\textbf{-1.83\%}) & 1.04h & \textbf{463.95} (\textbf{-2.55\%}) & 2.17h \\
\midrule
ACO I$_{1000}$ & 37.48 (3.28\%) & 2.03s & 97.00 (8.09\%) & 4.21s & 119.78 (11.53\%) & 7.32s & 303.58 (13.39\%) & 31.73s & 528.31 (10.96\%) & 1.15m \\
ACO I$_{2000}$ & 37.27 (2.71\%) & 4.04s & 95.88 (6.84\%) & 8.43s & 118.24 (10.10\%) & 14.71s & 299.46 (11.85\%) & 1.06m & 523.14 (9.88\%) & 2.30m \\
ACO I$_{5000}$ & 37.09 (2.20\%) & 10.09s & 94.73 (5.57\%) & 21.10s & 116.07 (8.08\%) & 35.87s & 294.24 (9.90\%) & 2.68m & 516.05 (8.39\%) & 5.77m \\
ACO I$_{10000}$ & 36.96 (1.85\%) & 20.17s & 93.76 (4.48\%) & 42.26s & 114.66 (6.76\%) & 1.24m & 290.70 (8.58\%) & 5.38m & 510.70 (7.26\%) & 11.55m \\
\midrule
DyNACO I$_{1000}$ & 37.30 (2.79\%) & 1.79s & 95.92 (6.89\%) & 4.28s & 119.26 (11.04\%) & 7.93s & 302.10 (12.84\%) & 34.73s & 525.91 (10.46\%) & 1.18m \\
DyNACO I$_{2000}$ & 37.11 (2.26\%) & 3.56s & 94.92 (5.78\%) & 8.46s & 117.90 (9.78\%) & 15.87s & 299.17 (11.74\%) & 1.16m & 521.58 (9.55\%) & 2.37m \\
DyNACO I$_{5000}$ & 36.80 (1.42\%) & 9.58s & 93.70 (4.41\%) & 21.37s & 115.37 (7.42\%) & 38.30s & 293.35 (9.57\%) & 2.84m & 513.55 (7.86\%) & 5.86m \\
DyNACO I$_{10000}$ & \textbf{36.67} (\textbf{1.04\%}) & 19.05s & 92.75 (3.36\%) & 42.87s & 113.89 (6.04\%) & 1.28m & 289.87 (8.27\%) & 5.60m & 508.23 (6.75\%) & 11.70m \\
\bottomrule
\end{tabular}
}
\end{table*}

We optimize using PPO with trajectory replay.
After collecting trajectories for $S$ sampling iterations under $\pi_{\theta_{\text{old}}}$, we perform $K_{\text{PPO}}$ optimization epochs using the clipped surrogate:
\begin{multline}
\mathcal{L}^{\text{PPO}}(\theta) = -\mathbb{E}_{a,s}\Big[ \min\big( r_{a,s}(\theta) \hat{A}_{a,s}, 
\text{clip}(r_{a,s}(\theta), 1{\pm}\epsilon) \hat{A}_{a,s} \big) \Big],
\end{multline}
where $r_{a,s}(\theta) = \pi_\theta(\pi_{a,s} | \mathcal{S}_h) / \pi_{\theta_{\text{old}}}(\pi_{a,s} | \mathcal{S}_h)$, $\epsilon=0.1$, and $\hat{A}_{a,s}$ is the normalized advantage.
We store pheromone snapshots and action traces for efficient probability ratio computation; the full procedure is given in Algorithm~\ref{alg:train}.

\subsection{Inference Strategy}
\label{sec:inference}
We use constant guidance ($\gamma{=}1$) during training, with no annealing or phased scheduling.
At inference, the guidance weight~$\gamma$ (Eq.~\ref{eq:held-logits}) can be modulated via two strategies: \emph{guidance annealing}, which linearly decays $\gamma \to 0$ across inner iterations; and \emph{phased injection}, which withholds neural guidance for the first fraction of outer steps.
The specific configuration varies by problem and scale: for TSP, we apply annealing at all scales and additionally enable phased injection at TSP-1K; for CVRP, we apply phased injection only at longer iteration budgets ($I{\geq}5000$) and never anneal.

\section{EXPERIMENTS}
\label{sec:experiments}

We evaluate \DyNACO{} on large-scale TSP and CVRP instances (1K--100K nodes), comparing against classical solvers, construction-based neural methods, and neural-ACO baselines.

\smallskip\noindent\textbf{Setups.}
All experiments use a fixed random seed and deterministic single-run evaluation. Training instances are generated online, with coordinates sampled uniformly from $[0,1]^2$, yielding 320 training instances in total (32 per epoch $\times$ 10 epochs). We train with PPO on a single RTX 5090 GPU and Ryzen 9 7950X CPU with 16 cores; default \DyNACO{} hyperparameters are summarized in Table~\ref{tab:hyperparams}, and baseline configurations are detailed in Appendix~\ref{app:setup}.

For evaluation, we largely follow prior works~\cite{attentionmodel,pomo,lehd,sil}. Instances at scales 1K, 5K, and 10K are taken from the benchmark datasets used in prior work~\cite{attentionmodel,pomo,lehd}, while 50K and 100K instances are generated by \cite{sil}. Each benchmark contains 128 instances at 1K and 16 instances at each larger scale; all reported objectives and gaps are averaged over the corresponding instance set. For CVRP, demands are sampled uniformly from ${1,\ldots,9}$ with capacity $Q$ scaled by problem size ($Q{=}50/250/500/1000/2000$ at 1K/5K/10K/50K/100K). The candidate graph is a symmetric $K$-NN graph with $K{=}32$, with ties broken by node index; a backup list of 64 additional neighbors is maintained for fallback~\cite{ACOTSP-MF}.

For real-world evaluation, following \cite{sil}, we extract all symmetric instances with Euclidean 2D coordinates (\texttt{EUC\_2D}) and more than 1K nodes from TSPLIB~\cite{tsplib} and CVRPlib~\cite{cvrplib}, yielding 33 TSP instances (1K--86K nodes) and 14 CVRP instances (1K--30K customers). These instances feature non-uniform, clustered, and geographically derived node distributions that differ substantially from the uniform training data.

\smallskip\noindent\textbf{Baselines.}
For comparison, we include Classical Solvers LKH-3~\cite{lkhtsp, lkhcvrp}; Construction-based NCO Methods POMO~\cite{pomo}, BQ~\cite{bq}, LEHD~\cite{lehd}, INViT~\cite{invit}, SIGD~\cite{sigd}, L2C-Insert~\cite{l2c}, SIL~\cite{sil}; Decomposition based Methods GLOP~\cite{glop}, H-TSP~\cite{htsp}, L2C-Insert I$_{1000}$~\cite{l2c} and SIL PRC$_{1000}$~\cite{sil}; Neural-ACO Methods: DeepACO~\cite{deepaco}, GFACS \cite{gfacs}, GTG-ACO~\cite{gtgaco} and HeatACO~\cite{heataco}; Unguided Backend (ACO): the perturbation-based ACO environment without neural guidance, which serves as the direct ablation baseline for DyNACO. For TSP, we adapt~\cite{faco}; for CVRP, we design a novel capacity-aware variant with SRR (Appendix~\ref{app:cvrp}).
All baselines are run on the same datasets, results taken from \cite{l2c, sil}. For baselines reported on other datasets (e.g., L2C-Insert~\cite{l2c}), we rerun them.

\smallskip\noindent\textbf{Metrics \& Inference.}
For comparison, we provide the average objective value (Obj.) and gap to reference (Gap). Obj.\ indicates the tour length, with shorter values indicating better performance. Gap measures the relative difference from reference solutions: for \TSP{}, these are LKH-3 outputs (provably optimal at 1K via Concorde verification; best-known at larger scales). For \CVRP{}, we use HGS~\cite{hgs} solutions as reference.
For our method, we present the results of the greedy search and different numbers of iterations.

\subsection{Comparative Results}
\label{sec:main-results}

Table~\ref{tab:main-results} presents results on synthetic \TSP{} and \CVRP{} instances.

\smallskip\noindent\textbf{DyNACO surpasses all neural baselines on TSP across all scales.}
\DyNACO{} achieves the lowest gap among all neural methods at every problem size, with the advantage widening at scale.
At TSP-100K, DyNACO delivers a 23\% relative quality improvement over the strongest competitor SIL at $55{\times}$ less runtime.
Most construction-based methods (POMO, LEHD, BQ, SIGD) run out of memory beyond 10K nodes; \DyNACO{} scales to 100K with sub-linear runtime growth.
Appendix~\ref{extended_ablation} reports peak GPU memory, including 3.11\,GB at TSP-100K versus OOM for dense baselines (Table~\ref{tab:memory-scaling}).
Performance also improves consistently with iteration budget, confirming that dynamic neural guidance compounds across outer steps rather than saturating early. Table~\ref{tab:neural-aco} further compares \DyNACO{} with prior neural-guided ACO methods at their maximum reported scales, showing that \DyNACO{} achieves better quality while substantially reducing runtime.

\smallskip\noindent\textbf{Dynamic neural guidance extends to CVRP with problem-standard features and backend logic.}
\DyNACO{} consistently improves upon the unguided ACO baseline at every iteration budget and problem scale on CVRP (Appendix~\ref{app:cvrp}), using the same semi-MDP training loop and state-aware representation. As in DeepACO, CVRP uses demand/capacity node features; DyNACO's added dynamic inputs are edge features based on pheromone and incumbent state, with CVRP-specific feasibility handled by the backend.
While SIL achieves strong synthetic-CVRP results at large scales, it requires substantially longer runtime.
More importantly, on real-world CVRPlib instances (Table~\ref{tab:realworld}), \DyNACO{} achieves a $2{\times}$ relative improvement over SIL, demonstrating that dynamic neural guidance generalizes to non-uniform distributions where SIL's synthetic advantage does not transfer.

\smallskip\noindent\textbf{Neural guidance improves quality while reducing runtime on TSP.}
Remarkably, on TSP, \DyNACO{} runs 23--33\% \emph{faster} than the unguided solver at matched iteration budgets, while simultaneously producing better solutions (Table~\ref{tab:time-breakdown}).
Better-targeted perturbations create less sub-optimality, so SRR's convergent repair terminates in fewer iterations; the neural policy amortizes its own inference cost by reducing downstream local-search work.
On CVRP, neural guidance adds less than 1\% to total wall-clock time at large scales while yielding consistent quality improvements.

\begin{table}[t]
\centering
\caption{Time breakdown (CPU vs.\ GPU) and quality for unguided ACO vs.\ DyNACO at $I_{10000}$.}
\label{tab:time-breakdown}
\small
\renewcommand{\arraystretch}{0.80}
\resizebox{\columnwidth}{!}{%
\begin{tabular}{llcccc}
\toprule
Problem & Method & Gap & CPU (s) & GPU (s) & Total (s) \\
\midrule
\multirow{2}{*}{\TSP{}1K}
  & ACO    & 0.48\% & 5.44  & --   & 5.44  \\
  & DyNACO & \textbf{0.20\%} & 3.52  & 0.16  & 3.68  \\
\midrule
\multirow{2}{*}{\TSP{}10K}
  & ACO    & 2.02\% & 26.81 & --   & 26.81 \\
  & DyNACO & \textbf{0.82\%} & 16.79 & 1.10  & 17.89 \\
\midrule
\multirow{2}{*}{\TSP{}100K}
  & ACO    & 3.12\% & 223.32 & --    & 223.32 \\
  & DyNACO & \textbf{1.90\%} & 159.85 & 11.84 & 171.69 \\
\midrule
\midrule
\multirow{2}{*}{\CVRP{}1K}
  & ACO    & 1.85\% & 20.17 & --   & 20.17 \\
  & DyNACO & \textbf{1.04\%} & 18.88 & 0.17  & 19.05 \\
\midrule
\multirow{2}{*}{\CVRP{}10K}
  & ACO    & 6.76\% & 74.43 & --   & 74.43 \\
  & DyNACO & \textbf{6.04\%} & 76.14 & 0.56  & 76.70 \\
\midrule
\multirow{2}{*}{\CVRP{}100K}
  & ACO    & 7.26\% & 692.81 & --   & 692.81 \\
  & DyNACO & \textbf{6.75\%} & 696.04 & 5.93  & 701.97 \\
\bottomrule
\end{tabular}}
\end{table}

\begin{table}[t]
\centering
\small
\renewcommand{\arraystretch}{0.85}
\caption{Comparison with neural-guided ACO methods at their maximum reported scale. HeatACO~\cite{heataco} combines ACO with other pretrained heatmap models ~\cite{attgcn, dimes, utsp, difusco}.}
\label{tab:neural-aco}
\resizebox{\columnwidth}{!}{%
\begin{tabular}{lcccccc}
\toprule
& \multicolumn{2}{c}{\TSP{}1K} & \multicolumn{2}{c}{\TSP{}10K} & \multicolumn{2}{c}{\CVRP{}1K} \\
\cmidrule(lr){2-3} \cmidrule(lr){4-5} \cmidrule(lr){6-7}
Method & Gap & Time & Gap & Time & Gap & Time \\
\midrule
DeepACO~\cite{deepaco}       & 2.87\% & 66s & -- & -- & 2.40\% & 1.3m \\
GFACS~\cite{gfacs}           & 2.63\% & 66s & -- & -- & 2.11\% & 1.3m \\
GTG-ACO~\cite{gtgaco}        & 2.42\% & -- & -- & -- & -- & -- \\
\midrule
ACO+AttGCN~\cite{attgcn}    & 0.44\% & 5s & 1.27\% & 1.47m & -- & -- \\
ACO+DIMES~\cite{dimes}     & 0.39\% & 6s & 1.15\% & 1.4m & -- & -- \\
ACO+UTSP~\cite{utsp}      & 0.42\% & 6s & -- & -- & -- & -- \\
ACO+DIFUSCO~\cite{difusco}   & 0.23\% & 10s & 1.19\% & 2.89m & -- & -- \\
\midrule
\textbf{DyNACO (ours)}                & \textbf{0.20\%} & 4s & \textbf{0.82\%} & 18s & \textbf{1.04\%} & 19s \\
\bottomrule
\end{tabular}}
\end{table}

\subsection{Training Efficiency}
\label{sec:runtime}

\DyNACO{} converges in approximately 30 minutes for TSP-1K and roughly 4 hours for TSP-100K on our machine, using only 320 training instances, compared to hours or days for SIL and LEHD (full training times and hardware details in Appendix~\ref{extended_ablation}).
The GNN encoder and MLP decoder contain 67,361 parameters, only 160 more than the corresponding static-input variant, and this count remains constant across problem scales because the architecture operates on local $K$-NN neighborhoods.
Appendix~\ref{extended_ablation} reports the static-vs-dynamic feature memory overhead in Table~\ref{tab:feature-memory-app}.

\subsection{Ablation Studies}
\label{sec:ablations}
\smallskip\noindent\textbf{Both \DyNACO{} mechanisms are essential; trajectory-aware training contributes to the larger effect.}
Table~\ref{tab:ablation-decoupling} isolates the two mechanisms of dynamic neural guidance.
Removing trajectory-aware training (reverting to static, single-step optimization) causes the largest degradation, while removing the state-aware representation also consistently hurts across all scales.
This confirms that both mechanisms are indispensable, and that aligning training with the iterative search dynamics is the more critical factor.

\begin{table}[t]
\centering
\caption{Decoupling the two mechanisms of dynamic neural guidance. We reported Gap (\%) of 16 instances to BKS at I$_{1000}$.}
\label{tab:ablation-decoupling}
\small
\resizebox{\columnwidth}{!}{%
\begin{tabular}{lcccc}
\toprule

Method & TSP1K $\downarrow$ & TSP5K $\downarrow$ & TSP10K $\downarrow$ & CVRP1K $\downarrow$ \\
\midrule
Static only       & 3.65 & 4.87 & 3.56 & 3.00 \\
+ Trajectory-aware Training   & 0.75 & 2.11 & 2.28 & 2.79 \\
+ State-aware Representation  (Full)     & \textbf{0.37} & \textbf{1.24} & \textbf{1.62} & \textbf{2.58} \\
\bottomrule
\end{tabular}
}
\end{table}

\smallskip\noindent\textbf{SRR preserves both training signal and search quality.} We discussed in Figure~\ref{fig:credit_assignment} that: full local search (FLS) can produce strong individual solutions, but it also rewrites the sampled perturbation enough to wash out the model's contribution and create negative training progress; truncated local search (TLS), used by previous neural-guided ACO methods \cite{deepaco,gfacs}, limits operators to $N/4$ and preserves more learning signal, but leaves tours under-refined.
In contrast, SRR runs to convergence within the perturbation scope, preserving the causal link between the policy action and the post-refinement reward while retaining a low refinement cost. Table~\ref{tab:srr-training} further confirms this training effect: SRR yields the largest improvement over the untrained model, whereas FLS produces negative progress due to gradient washout. At inference, Table~\ref{tab:srr-inference} shows that SRR also gives the best quality-runtime trade-off, outperforming both FLS and TLS on TSP-5K.

\begin{table}[t]
\centering
\caption{Training progress under different refinement strategies. Values are improvement over the untrained model.}
\label{tab:srr-training}
\small
\setlength{\tabcolsep}{0pt}
\renewcommand{\arraystretch}{0.80}
\begin{tabular*}{\columnwidth}{@{\extracolsep{\fill}}lccc@{}}
\toprule
Epoch & SRR $\uparrow$ & TLS $\uparrow$ & FLS $\uparrow$ \\
\midrule
0 (untrained) & 0.000\% & 0.000\% & 0.000\% \\
1 & \textbf{+0.224\%} & +0.099\% & -0.003\% \\
5 & \textbf{+0.432\%} & +0.146\% & -0.016\% \\
9 & \textbf{+0.438\%} & +0.127\% & -0.023\% \\
\midrule
Avg. ACO time & \textbf{2.35s} & 4.33s & 8.03s \\
\bottomrule
\end{tabular*}
\end{table}

\begin{table}[t]
\centering
\caption{Inference quality and runtime under different refinement strategies (TSP-5K, I$_{1000}$).}
\label{tab:srr-inference}
\small
\renewcommand{\arraystretch}{0.85}
\begin{tabular*}{\columnwidth}{@{\extracolsep{\fill}}lccc@{}}
\toprule
Refinement & Time (s) & Obj. & Gap to LKH-3 \\
\midrule
FLS & 13.63 & 53.76 & 5.47\% \\
TLS & 7.62 & 57.80 & 13.39\% \\
SRR & \textbf{1.93} & \textbf{51.66} & \textbf{1.35\%} \\
\bottomrule
\end{tabular*}
\end{table}

\smallskip\noindent\textbf{Pheromone stabilization improves consistency without replacing dynamic guidance.}
Pheromone stabilization improves the consistency of the training signal for both the unguided backend and \DyNACO{}.
Table~\ref{tab:ps-ablation} shows that dynamic guidance remains beneficial with or without stabilized pheromone bounds, while stabilization further improves the final guided solver.

\begin{table}[t]
\centering
\caption{Ablation of pheromone stabilization (PS). We reported Gap (\%) to BKS.}
\label{tab:ps-ablation}
\small
\renewcommand{\arraystretch}{0.85}
\begin{tabular*}{\columnwidth}{@{\extracolsep{\fill}}lcccc@{}}
\toprule
Dataset & ACO w/o PS $\downarrow$ & \DyNACO{} w/o PS $\downarrow$ & ACO $\downarrow$ & \DyNACO{} $\downarrow$ \\
\midrule
CVRP-1K & 4.09 & \textbf{3.51} & 2.94 & \textbf{2.50} \\
CVRPlib (5) & 2.91 & \textbf{2.71} & 2.76 & \textbf{2.18} \\
\bottomrule
\end{tabular*}
\end{table}

\smallskip\noindent\textbf{Inference-time strategies provide complementary gains on TSP.}
All models are trained identically on the full trajectory with constant $\gamma{=}1$.
At inference, guidance annealing and phased injection are applied as problem and scale-specific hyperparameters (Table~\ref{tab:inference-config}).
On TSP, both strategies yield consistent improvements; on CVRP, only phased injection at long horizons is beneficial (Table~\ref{tab:ablation-inference}).
Appendix~\ref{extended_ablation} further ablates inference strategies and reports sensitivity to $K$, $M$, and the temporal abstraction granularity $(H,S)$.

\begin{table}[t]
\centering
\caption{Inference strategy configuration per problem \& scale.}
\label{tab:inference-config}
\small
\renewcommand{\arraystretch}{0.85}
\begin{tabular*}{\columnwidth}{@{\extracolsep{\fill}}lcc@{}}
\toprule
Setting & Annealing & Phased Injection \\
\midrule
TSP-1K (all budgets)          & \checkmark & \checkmark \\
TSP-5K, 10K, 50K, 100K       & \checkmark &            \\
CVRP (all scales, $I{<}5000$) &            &            \\
CVRP (all scales, $I{\geq}5000$) &         & \checkmark \\
\bottomrule
\end{tabular*}
\end{table}

\begin{table}[h]
\centering
\caption{Inference strategy ablation with gap (\%) to BKS. All models trained with $\gamma{=}1$; strategies applied post-hoc.}
\label{tab:ablation-inference}
\small
\begin{tabular*}{\columnwidth}{@{\extracolsep{\fill}}lcccc@{}}
\toprule
Inference Strategy & TSP5K $\downarrow$ & TSP10K $\downarrow$ & CVRP5K $\downarrow$ & CVRP10K $\downarrow$ \\
\midrule
Constant ($\gamma{=}1$)            & 1.38 & 1.73 & \textbf{6.58} & 11.16 \\
Phased injection           & \textbf{1.20} & 1.68 & 7.07 & \textbf{10.86} \\
Guidance annealing         & 1.23 & \textbf{1.59} & 6.84 & 11.17 \\
Phased + Annealing         & 1.23 & 1.60 & 7.35 & 11.02 \\
\bottomrule
\end{tabular*}
\end{table}

\subsection{Extension to Other COP}
Table~\ref{tab:other-cops} shows results on Orienteering, Multiple Knapsack, and Bin Packing, using the same environment as \cite{deepaco}, applied with trajectory-aware training.
These runs use the same semi-MDP dynamic-guidance formulation without SRR, isolating whether the dynamic policy interface can improve existing ACO algorithms beyond TSP and CVRP.
Dynamic guidance improves over both unguided ACO and static guidance on all three problems, suggesting that the framework is not tied to a routing-specific objective, even though SRR is still necessary as a scalability mechanism in large-scale routing problems.

\begin{table}[t]
\centering
\caption{Extension to additional COPs. Higher is better.}
\label{tab:other-cops}
\small
\renewcommand{\arraystretch}{0.85}
\begin{tabular*}{\columnwidth}{@{\extracolsep{\fill}}lccc@{}}
\toprule
Method & OP $\uparrow$ & MKP $\uparrow$ & BPP $\uparrow$ \\
\midrule
ACO & 7.9222 & 24.1082 & 0.8608 \\
Static & 8.4071 & 24.1426 & 0.9139 \\
Dynamic & \textbf{10.4177} & \textbf{24.1477} & \textbf{0.9205} \\
\bottomrule
\end{tabular*}
\end{table}

\subsection{Interpretation and Analysis}
\label{sec:analysis}

We now examine what dynamic neural guidance learns that static heatmaps fundamentally cannot capture.

\smallskip\noindent\textbf{The policy adapts its guidance to the current search phase.}
Standard neural-ACO methods produce a static heatmap that is identical whether the solver has run for 10 or 10,000 iterations.
Because \DyNACO{} conditions on the evolving pheromone field and incumbent, improvements grow with longer search horizons: a model trained with $I{=}1000$ total iterations yields \emph{larger} gains when run for $I{=}10000$ (Table~\ref{tab:main-results}).
A static heatmap cannot exhibit this behavior, confirming that state-aware conditioning is what enables the policy to remain useful across all search phases.

\begin{figure}[t]
    \centering
    \includegraphics[width=\columnwidth]{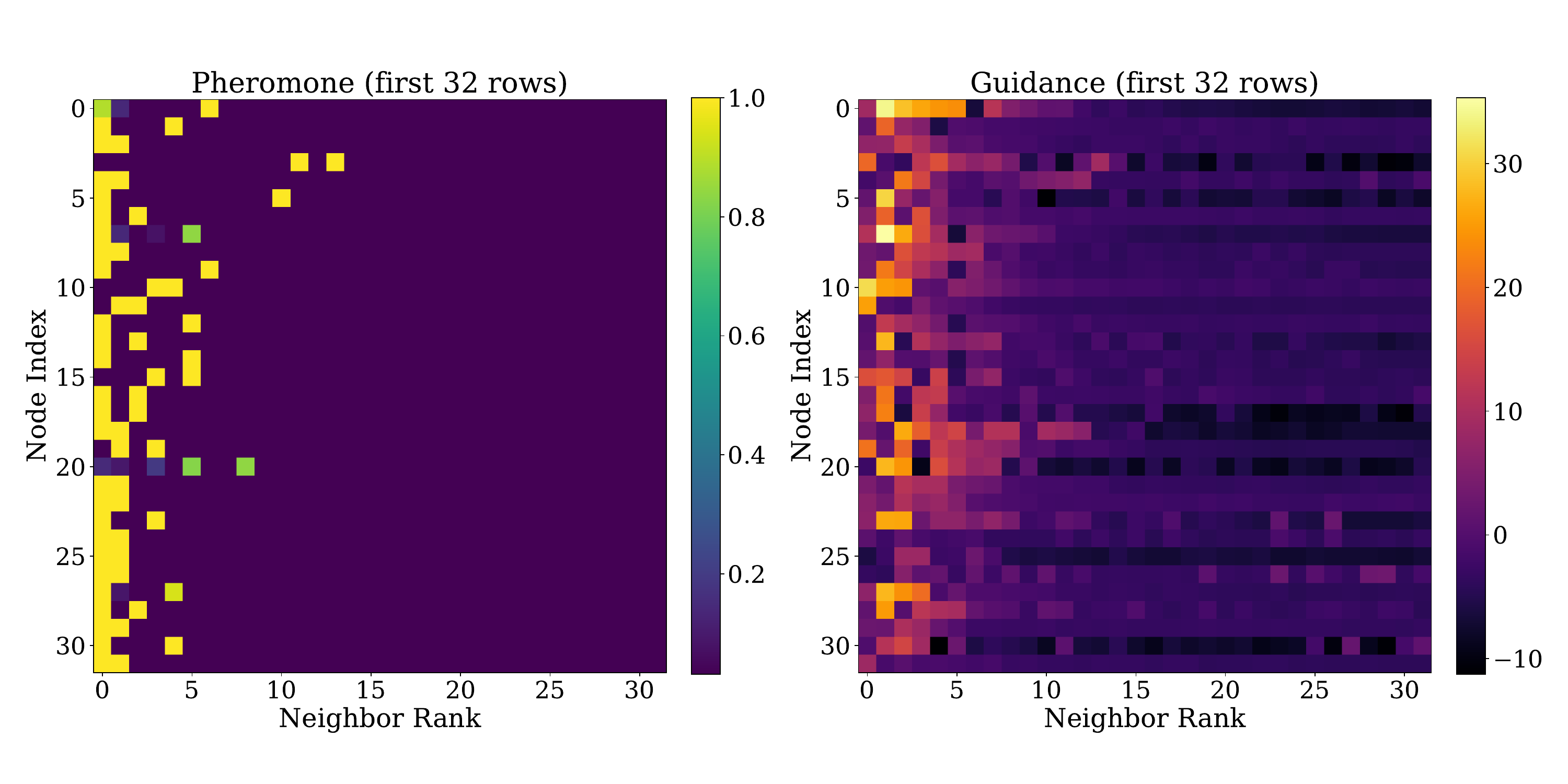}
    \caption{Visualization of pheromone concentration and neural guidance scores (first 32 rows of the edge matrix) at successive iterations~$T$.}
    \label{fig:matrix_model}
\end{figure}

\smallskip\noindent\textbf{The policy learns to counteract ACO stagnation.}
Figure~\ref{fig:matrix_model} displays the pheromone matrix and corresponding neural guidance scores (first 32 rows) at successive iterations.
As the search progresses, pheromone concentrations saturate: most edges converge to $\tau_{\max}$, leaving the pheromone signal nearly uninformative.
Despite this uniformity, the model outputs a diverse range of guidance scores.
Most notably, edges at maximum pheromone concentration almost always receive \emph{negative} guidance, actively suppressing over-reinforced edges and redirecting exploration.
This selective counter-pressure is unavailable to static priors, which lack access to the pheromone landscape, confirming that dynamic neural guidance learns qualitatively different anti-stagnation strategies.

\smallskip\noindent\textbf{The 1K-trained model transfers zero-shot to larger scales with minimal degradation.}
Table~\ref{tab:cross-scale} evaluates the 1K model on 5K and 10K instances without retraining.
Remarkably, on TSP-5K the 1K model \emph{outperforms} the in-scale model, while on TSP-10K and CVRP the degradation is negligible.
This transferability stems from three scale-invariant design choices: the local $K$-NN architecture whose neighborhood statistics are size-independent, the state-aware representation that uses normalized features abstracting away absolute magnitudes, and the perturbation-based action space whose fixed horizon ($M{=}12$) does not grow with $N$ unlike constructive methods.

\begin{table}[t]
\centering
\caption{Cross-scale zero-shot transfer: 1K model evaluated on larger instances.}
\label{tab:cross-scale}
\small
\renewcommand{\arraystretch}{0.80}
\begin{tabular*}{\columnwidth}{@{\extracolsep{\fill}}llcccc@{}}
\toprule
& & \multicolumn{2}{c}{$I_{1000}$} & \multicolumn{2}{c}{$I_{10000}$} \\
\cmidrule(lr){3-4} \cmidrule(lr){5-6}
Problem & Model & Gap & $\Delta$ & Gap & $\Delta$ \\
\midrule
\multirow{2}{*}{\TSP{}5K}
  & In-scale  & 1.30\% & --       & 0.75\% & --       \\
  & 1K model  & \textbf{1.23\%} & $-$0.07  & \textbf{0.72\%} & $-$0.03  \\
\midrule
\multirow{2}{*}{\TSP{}10K}
  & In-scale  & \textbf{1.59\%} & --       & \textbf{0.82\%} & --       \\
  & 1K model  & 1.66\% & +0.07    & 0.90\% & +0.08    \\
\midrule
\multirow{2}{*}{\CVRP{}5K}
  & In-scale  & \textbf{6.89\%} & --       & \textbf{3.36\%} & --       \\
  & 1K model  & 7.10\% & +0.21    & 3.60\% & +0.24    \\
\midrule
\multirow{2}{*}{\CVRP{}10K}
  & In-scale  & \textbf{11.04\%} & --      & \textbf{6.04\%} & --       \\
  & 1K model  & 11.72\% & +0.68   & 6.07\% & +0.03    \\
\bottomrule
\end{tabular*}
\end{table}

\smallskip\noindent\textbf{DyNACO generalizes zero-shot to real-world benchmarks across both problems.}
We evaluate the 1K-trained model on 33 TSPLIB instances (1K--86K nodes) and 14 CVRPlib instances (1K--30K customers) without any fine-tuning (Table~\ref{tab:realworld}).
While many neural baselines OOM beyond $\sim$10K nodes, DyNACO solves \emph{all} instances.
On TSPLIB, DyNACO achieves a 31\% relative improvement over unguided ACO, winning on 29/33 instances; on CVRPlib, it improves every instance with a 14\% relative reduction in gap.
These results confirm that dynamic neural guidance trained only on uniform synthetic 1K instances can transfer robustly to non-uniform distributions, real-world topologies, and scales up to instances $86{\times}$ larger than the training ones.

\begin{table}[t]
\centering
\caption{Real-world benchmark results (zero-shot from 1K model, $I_{10000}$).}
\label{tab:realworld}
\small
\renewcommand{\arraystretch}{0.80}
\begin{tabular*}{\columnwidth}{@{\extracolsep{\fill}}lcccc@{}}
\toprule
& \multicolumn{2}{c}{TSPLIB (33)} & \multicolumn{2}{c}{CVRPlib (14)} \\
\cmidrule(lr){2-3} \cmidrule(lr){4-5}
Method & Gap (\%) & Time & Gap (\%) & Time \\
\midrule
LKH3 & 0.07 & 35m & 13.6 & 2.1h \\
HGS & -- & -- & 5.15 & 5h \\
\midrule
LEHD~\cite{lehd}  & 13.2 & 38m & 16.9 & 40m \\
GLOP~\cite{glop} & 6.99 & 34s & 20.6 & 39s \\
SIL~\cite{sil} & 3.03 & 45m & 7.69 & 54m \\
\midrule
\textbf{DyNACO (ours)}       & \textbf{0.89} & 11.57s & \textbf{3.66} & 51.07s \\
\bottomrule
\end{tabular*}
\end{table}

\begin{figure}[t]
    \centering

    \subfigure[Training correlation\label{fig:guidance_eta_corr}]{%
        \includegraphics[width=.49\columnwidth]{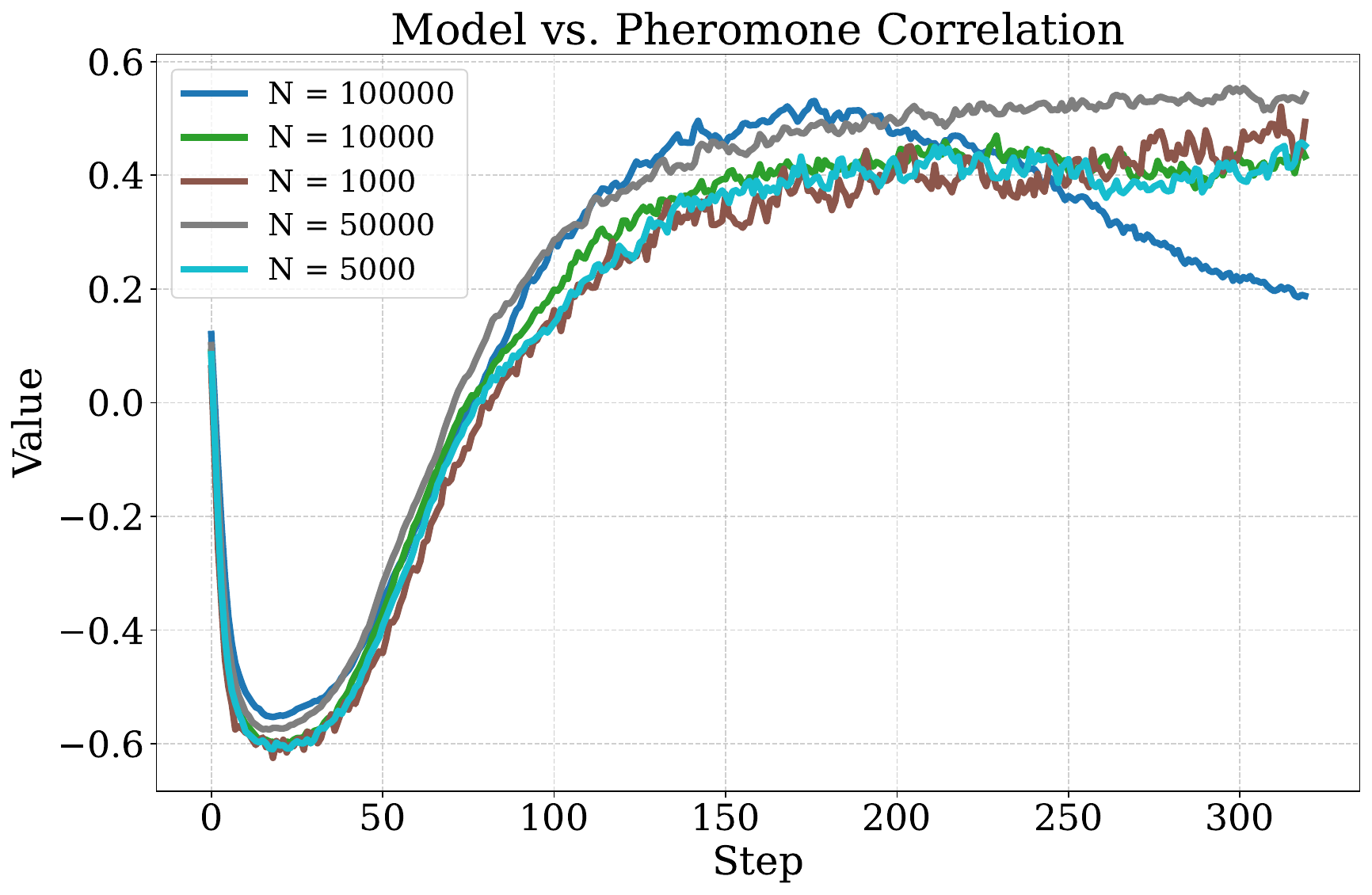}%
    }
    \subfigure[Testing Interaction\label{fig:interaction}]{%
        \includegraphics[width=.49\columnwidth]{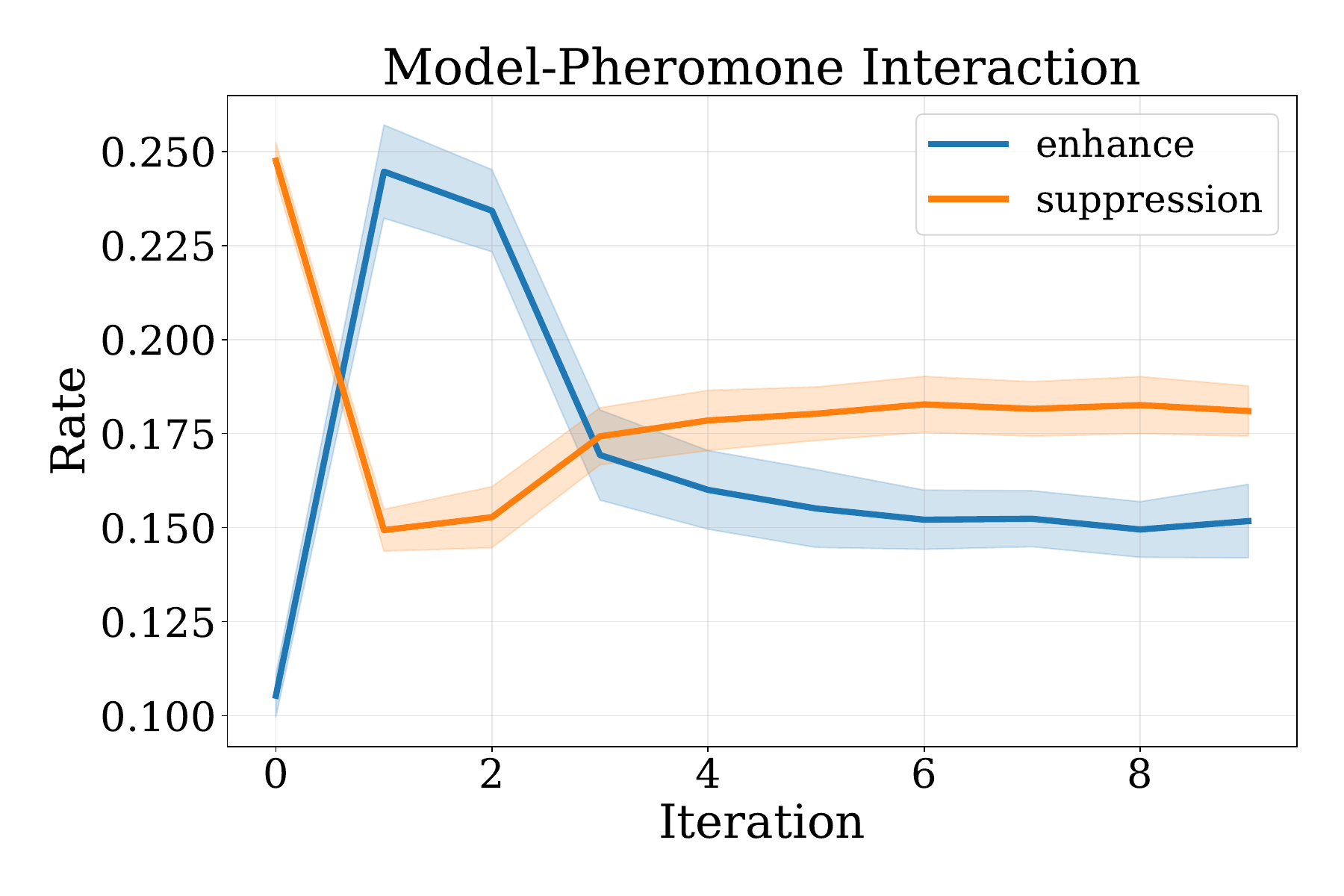}%
    }\hfill

    \caption{Visualization of the guidance-pheromone dynamics during both training and inference.}
    \label{fig:guidance_pheromone_dynamics}
\end{figure}

\smallskip\noindent\textbf{Cross-scale guidance-pheromone dynamics.} Figure~\ref{fig:guidance_pheromone_dynamics} shows that the learned policy exhibits a consistent two-phase interaction with pheromone across scales and search steps. In training, the guidance-pheromone correlation follows the same trajectory for $N{=}1$K--$100$K: an early sharp decrease, indicating that the policy first learns to diverge from pheromone and provide complementary guidance, followed by gradual recovery as it selectively re-aligns with useful pheromone signals. In testing, along the search trajectory, the interaction shifts from cooperation to opposition: at the outer step $H{=}1$, enhancement is high (${\sim}0.24$) and suppression is low (${\sim}0.15$), but by $H{=}3$ suppression overtakes enhancement and remains dominant through $H{=}10$. This crossover reflects a learned adaptation: early in search, when pheromone is weakly informative, the policy amplifies promising signals; later, as pheromone concentrates and risks premature convergence, it suppresses over-reinforced edges to preserve diversity. These results collectively demonstrate that trajectory-aware training can achieve an adaptive cooperative-to-adversarial guidance strategy, which static priors cannot express due to a lack of access to constantly changing pheromone states.

\section{CONCLUSION}

This work introduces DyNACO, a framework that formalizes the neural-ACO interaction as a semi-MDP and uses a lightweight meta-policy to inject dynamic guidance throughout the search trajectory. DyNACO combines two mechanisms, state-aware representation and trajectory-aware training, with scope-restricted refinement to maintain stable policy-reward credit assignment while scaling to 100K-node instances. Extensive results demonstrate state-of-the-art performance among neural methods on TSP and CVRP across all evaluated scales, with up to 59\% relative gap reduction over the unguided ACO baseline (TSP-10K, I$_{10000}$), and a 23--33\% runtime reduction on TSP. Trained only on uniform synthetic 1K instances, DyNACO generalizes zero-shot to real-world benchmarks, surpassing all neural baselines on both TSPLIB and CVRPlib and outperforming the classical solver HGS on CVRPlib.

One limitation is that the temporal abstraction granularity $(H,S)$ and the $K$-NN candidate graph require problem-specific configuration, and models are currently trained at a fixed scale.
Future work includes: (1)~adaptive candidate selection to address the fixed-$K$ reachability constraint; (2)~adaptive temporal abstraction, e.g., changing the macro-action duration $S$ based on stagnation while preserving the semi-MDP abstraction; (3)~curriculum or multi-scale training for improved cross-scale generalization, which must handle feature-scale discrepancies and possible gradient conflicts across instance sizes and may require meta-learning or scale-normalized attention designs; and (4)~extending dynamic neural guidance to other combinatorial domains and meta-heuristic families.


\bibliographystyle{ACM-Reference-Format}
\balance
\bibliography{references}

\newpage
\appendix

\setcounter{table}{0}
\renewcommand{\thetable}{A\arabic{table}}
\setcounter{figure}{0}
\renewcommand{\thefigure}{A\arabic{figure}}
\setcounter{algorithm}{0}
\renewcommand{\thealgorithm}{A\arabic{algorithm}}

\section{State-Aware Representation Features}
\label{app:state-features}

\DyNACO{} keeps the static node features used by DeepACO~\cite{deepaco} and adds six sparse edge features for each candidate edge $(i,j)$ in the $K$-NN graph. Let
$\bar d_i=K^{-1}\sum_{j'\in\mathcal{N}_K(i)}d_{ij'}$ and
$\bar\tau_i=K^{-1}\sum_{j'}\tau_{ij'}$. The added features are summarized in Table~\ref{tab:state-features}; pheromone features are clamped to fixed ranges for numerical stability.

\begin{table}[h]
\centering
\caption{Dynamic edge features used by the state-aware policy.}
\label{tab:state-features}
\small
\setlength{\tabcolsep}{3pt}
\renewcommand{\arraystretch}{0.95}
\begin{tabular*}{\columnwidth}{@{}c@{\extracolsep{\fill}}p{0.39\columnwidth}p{0.42\columnwidth}@{}}
\toprule
Feature & Definition & Meaning \\
\midrule
$f_1$ & $d_{ij}/\bar d_i$ & Scale-normalized distance. \\
$f_2$ & $\sigma_{\tau_i}/\bar\tau_i$ & Local pheromone dispersion. \\
$f_3$ & $\log(\tau_{ij}/\bar\tau_i)$ & Edge pheromone relative to other candidates of $i$. \\
$f_4$ & $\mathbf{1}[j=\mathrm{succ}_b(i)]$ & Incumbent successor edge. \\
$f_5$ & $\mathbf{1}[j=\mathrm{pred}_b(i)]$ & Incumbent predecessor edge. \\
$f_6$ & $\mathbf{1}[j\notin\{\mathrm{succ}_b(i),\mathrm{pred}_b(i)\}]$ & Edge that introduces new tour structure. \\
\bottomrule
\end{tabular*}
\end{table}

For static node features, we follow \cite{deepaco}, with node coordinates for \TSP{} and demand/capacity for \CVRP{}. The dynamic features are recomputed at every outer step from the current pheromone field and incumbent solution. With stabilized pheromone bounds $\tau_{\max}=1$ and $\tau_{\min}=1/K$, the representation has six edge channels and $O(nK)$ size for fixed $K$.

\section{Environment and Backend}
\label{app:backend}

This section provides a detailed description of the perturbation-based ACO environment (Section \ref{sec:backend}), covering the ant construction procedure, scope-restricted refinement, and CVRP-specific capacity handling.

\subsection{Perturbation Environment}
\label{app:perturbation}

For \TSP{}, we follow~\cite{faco}, ants perform $M$ relocation steps starting from the incumbent $b$, where node selection follows the neurally-shaped kernel $p^\theta$. Each ant touches at most $O(M \cdot K)$ edges, where $M \ll N$ and $K{=}32$, independent of problem size.

\subsubsection{CVRP Algorithms}
\label{app:cvrp}

For \CVRP{}, we design a novel pertubation-based algorithm. We maintain an explicit multi-route structure using linked lists, where each route $r$ tracks its current load $L_r$.
Transitions to node $j$ are masked if $\text{Load} + d_j > Q$ (capacity exceeded).
Specifically:
\begin{enumerate}
    \item \textbf{Same-route} ($r_u = r_v$): Always feasible; reordering within a route preserves load.
    \item \textbf{Cross-route} ($r_u \neq r_v$): Feasible only if $L_{r_u} + d_v \leq Q$.
    \item \textbf{Depot target} ($v = 0$): Triggers a route split; always feasible.
\end{enumerate}
The new-edge counter tracks cross-route edges specifically, ensuring sufficient inter-route perturbation for effective local search. The policy follows the DeepACO feature convention for static node inputs, including demand/capacity features. DyNACO adds the same dynamic edge-state features as in TSP, derived from pheromone and incumbent information. CVRP-specific feasibility logic is handled by the backend, which masks infeasible transitions and restricts local-search moves to capacity-preserving operations.

\subsubsection{Scope-Restricted CVRP Local Search}
The LS phase applies four operators restricted to the checklist of perturbed nodes:

\textbf{Intra-Route 2-opt.}
For each checklist node $u$, we examine 2-opt moves within $u$'s route, scanning $K$-nearest same-route neighbors.

\textbf{Inter-Route Relocate.}
For cross-route pairs $(u, v)$ with $u$ in the checklist, we evaluate relocating $u$ adjacent to $v$, subject to $L_{r_v} + d_u \leq Q$.

\textbf{Inter-Route Swap.}
We evaluate swapping positions of $u$ and $v$ between routes, requiring $L_{r_u} - d_u + d_v \leq Q$ and $L_{r_v} - d_v + d_u \leq Q$.

\textbf{2-opt* (Cross-Route Segment Exchange).}
We evaluate exchanging route tails, checking cumulative demands of resulting segments.

All operators use \emph{don't-look bit} (DLB) optimization: nodes are deactivated when no improving move is found and reactivated only when a neighbor participates in an accepted move.

\subsection{Scope-Restricted Refinement (SRR)}
\label{app:credit}

\subsubsection{The ``BFS Repair'' Mechanism}
\DyNACO{} exploits the locality of perturbation-based search. Let $\mathcal{B}^*$ be a 2-opt optimal incumbent. A policy perturbation replaces edges $E_{\mathrm{rem}}$ with $E_{\mathrm{add}}$, yielding $\mathcal{B}'$. Since only the neighborhood of $E_{\mathrm{add}}$ can become suboptimal, SRR initializes a checklist with the endpoints of the added edges and propagates improving moves outward until no checked node admits improvement.

\smallskip\noindent\textbf{Optimality.}
If the proposed SRR terminates with an empty checklist, the resulting tour $\mathcal{B}''$ is called 2-opt optimal under the candidate-graph move set.

\begin{proof}
Let $\mathcal{C}$ be the set of edges modified by the perturbation or SRR. All edges outside $\mathcal{C}$ are unchanged from the 2-opt optimal incumbent $\mathcal{B}^*$, so no improving 2-opt move can use only those edges. SRR starts from all perturbed endpoints and reactivates affected nodes after each accepted move; hence an empty checklist certifies that no candidate-graph 2-opt move involving any edge in $\mathcal{C}$ is improving. Cross-boundary moves also involve an edge in $\mathcal{C}$ and are therefore covered. Thus no improving candidate-graph 2-opt move remains in $\mathcal{B}''$.
\end{proof}

\begin{table}[b]
\centering
\caption{Default hyperparameters for \DyNACO{}.}
\label{tab:hyperparams}
\small
\setlength{\tabcolsep}{4pt}
\renewcommand{\arraystretch}{0.93}
\begin{tabular*}{\columnwidth}{@{}l@{\extracolsep{\fill}}p{0.66\columnwidth}@{}}
\toprule
Component & Setting \\
\midrule
ACO & $m=100$, $H=10$, $S=100$, $M=12$ \\
Pheromone & $\rho=0.1/0.5$ (\TSP{}/\CVRP{}), $\tau_{\max}=1$, $\tau_{\min}=1/K$ \\
Graph & $K=32$ candidates, 32 backup neighbors \\
PPO training & lr $=5{\times}10^{-6}$, clip $\epsilon=0.1$, $K_{\text{PPO}}=4$, AdamW+cosine, $\gamma=1$ \\
Architecture & 12 GNN layers, hidden dim 32, 3-layer MLP \\
\bottomrule
\end{tabular*}
\end{table}

\begin{table}[t]
\centering
\caption{Environment stabilization. We report results without inference strategies.}
\label{tab:ablation-mmas-app}
\small
\renewcommand{\arraystretch}{0.85}
\begin{tabular*}{\columnwidth}{@{\extracolsep{\fill}}lccc@{}}
\toprule
Bound Strategy & Obj. & Gap (\%)\\
\midrule
ACO (Standard \MMAS{}) & 51.8318 & 1.68  \\
ACO (Stabilized) & 52.0747 & 2.16 \\
DyNACO (Standard \MMAS{}) & 51.7022 & 1.43 \\
DyNACO (Stabilized) & 51.6795 & 1.38 \\
\bottomrule
\end{tabular*}
\end{table}

\begin{table}[t]
\centering
\caption{Sensitivity to temporal abstraction granularity: the number of outer guidance updates $H$ and inner ACO steps $S$. Each cell reports objective / time (s).}
\label{tab:hs-sensitivity-app}
\footnotesize
\setlength{\tabcolsep}{2.5pt}
\renewcommand{\arraystretch}{0.9}
\begin{tabular*}{\columnwidth}{@{\extracolsep{\fill}}lccc@{}}
\toprule
Dataset & $H/S=5/200$ & $10/100$ & $20/50$ \\
\midrule
\TSP{}-1K  & 23.22 / 0.32 & 23.23 / \textbf{0.32} & \textbf{23.21} / 0.42 \\
\TSP{}-5K  & 51.59 / \textbf{0.75} & 51.63 / 0.76 & \textbf{51.56} / 0.87 \\
\CVRP{}-1K & 36.62 / 1.34 & \textbf{36.28} / \textbf{1.27} & 36.49 / 1.33 \\
\CVRP{}-5K & \textbf{96.16} / 3.07 & 96.81 / \textbf{2.73} & 97.17 / 2.82 \\
\bottomrule
\end{tabular*}
\end{table}

\begin{table}[t]
\centering
\caption{Peak GPU memory usage (GB) on TSP instances. OOM: out of memory.}
\label{tab:memory-scaling}
\small
\renewcommand{\arraystretch}{0.85}
\begin{tabular*}{\columnwidth}{@{\extracolsep{\fill}}lccccc@{}}
\toprule
Method & 1K & 5K & 10K & 50K & 100K \\
\midrule
POMO & 0.10 & 2.54 & 10.11 & OOM & OOM \\
SIGD & 0.04 & 0.95 & 3.76 & OOM & OOM \\
LEHD & 0.10 & 2.27 & 9.01 & OOM & OOM \\
DeepACO & 0.14 & 2.11 & 8.31 & OOM & OOM \\
\DyNACO{} & \textbf{0.08} & \textbf{0.18} & \textbf{0.35} & 1.57 & 3.11 \\
\bottomrule
\end{tabular*}
\end{table}

\section{Experimental Setup}
\label{app:setup}

We use a 12-layer GNN encoder~\cite{deepaco} with residual connections and batch normalization; the decoder is a 3-layer MLP projecting edge embeddings to scalar log-priors. Default training and backend hyperparameters are summarized in Table~\ref{tab:hyperparams}.

\subsection{Extended Ablation Tables}
\label{extended_ablation}

\subsubsection{Environment Stabilization}

Table~\ref{tab:ablation-mmas-app} examines pheromone bound strategies.
Standard \MMAS{} couples bounds to solution cost, creating non-stationary dynamics.
Our stabilized variant ($\tau_{\max}{=}1$, $\tau_{\min}{=}1/K$) yields scale-invariant state representations and reduced training variance.

\subsubsection{Memory Scaling}

Table~\ref{tab:memory-scaling} reports peak GPU memory usage for representative constructive and neural-ACO baselines.
Methods based on dense attention or dense heatmaps scale quadratically in the number of nodes and become infeasible at larger scales.
\DyNACO{} instead operates on a fixed-size sparse candidate graph and uses perturbation-based sampling, so memory scales approximately linearly with $n$ for fixed $K$.

\subsubsection{Hyperparameter Sensitivity}

Table~\ref{tab:k-sensitivity-app} and Table~\ref{tab:m-sensitivity-app} evaluate the candidate-graph size $K$ and perturbation length $M$.
Larger $K$ improves quality with a near-linear runtime trade-off.
The best $M$ is problem-dependent: more perturbation steps are not automatically better, because excessive changes can disrupt the incumbent basin and increase refinement cost.

\begin{table}[t]
\centering
\caption{Sensitivity to candidate-graph size $K$.}
\label{tab:k-sensitivity-app}
\small
\renewcommand{\arraystretch}{0.85}
\begin{tabular*}{\columnwidth}{@{\extracolsep{\fill}}lcccc@{}}
\toprule
$K$ & TSP-5K & Time & CVRP-5K & Time \\
\midrule
32 & 51.63 & \textbf{0.76} & 96.81 & \textbf{2.73} \\
48 & 51.49 & 0.98 & 95.83 & 3.51 \\
64 & \textbf{51.47} & 1.17 & \textbf{95.08} & 4.42 \\
\bottomrule
\end{tabular*}
\end{table}

\begin{table}[t]
\centering
\caption{Sensitivity to perturbation length $M$.}
\label{tab:m-sensitivity-app}
\small
\renewcommand{\arraystretch}{0.85}
\begin{tabular*}{\columnwidth}{@{\extracolsep{\fill}}lcccc@{}}
\toprule
$M$ & TSP-5K & Time & CVRP-5K & Time \\
\midrule
8 & \textbf{51.62} & \textbf{0.73} & 97.13 & \textbf{2.37} \\
12 & 51.63 & 0.76 & 96.81 & 2.73 \\
16 & 51.77 & 0.82 & \textbf{96.48} & 3.24 \\
\bottomrule
\end{tabular*}
\end{table}

Table~\ref{tab:hs-sensitivity-app} varies the outer guidance updates $H$ and inner ACO steps $S$ while keeping the total iteration budget fixed.
\DyNACO{} remains robust across these temporal granularities, supporting the fixed-duration semi-MDP abstraction used in the main method.




\subsubsection{Parameter and Feature Overhead}

\DyNACO{} adds dynamic state features to a compact GNN policy.
Using the architecture of Ye et al.~\cite{deepaco}, increasing the edge-feature input from 1 to 6 adds only 160 parameters, from 67,201 to 67,361 (Table~\ref{tab:feature-memory-app}).
The peak GPU-memory overhead of these additional features remains small even at 100K nodes, as shown in Table~\ref{tab:feature-memory-app}.

\begin{table}[t]
\centering
\caption{Peak GPU memory (GB) for static and dynamic policy inputs.}
\label{tab:feature-memory-app}
\small
\renewcommand{\arraystretch}{0.85}
\begin{tabular*}{\columnwidth}{@{\extracolsep{\fill}}llcc@{}}
\toprule
Problem & Scale & Static (GB) & Dynamic (GB) \\
\midrule
TSP & 10K & 0.286 & 0.293 \\
TSP & 100K & 2.524 & 2.584 \\
CVRP & 10K & 0.286 & 0.292 \\
CVRP & 100K & 2.513 & 2.573 \\
\bottomrule
\end{tabular*}
\end{table}

\section{Scalability of Other Neural Baselines}
\label{app:related}

End-to-end constructive policies~\cite{attentionmodel,pomo,bq,lehd, invit, sigd} are typically trained on small-scale instances ($N{=}50$ or $100$) and rely on quadratic attention complexity $O(N^2)$ in the encoder.
When applied to large-scale instances ($N{\geq}1{,}000$), these models suffer from out-of-distribution degradation, as the learned representations do not transfer across scales, compounded by memory exhaustion from the $O(N^2)$ encoder.
The ``OOM'' entries in Table~\ref{tab:main-results} reflect this inherent limitation.
Prior neural-ACO methods have the same dense bottleneck in a different form: they materialize heatmap or pheromone-prior matrices in $\mathbb{R}^{N\times N}$ and are typically evaluated only up to $N{\leq}1000$, so scaling the full ACO loop induces an $O(N^2 m T)$ memory/computation burden across ants and iterations.

\smallskip\noindent\textbf{Decomposition attempts.}
Recent methods address this through divide-and-conquer (GLOP~\cite{glop}, H-TSP~\cite{htsp}) or hierarchical construction (SIL~\cite{sil}, LEHD~\cite{lehd}).
However, decomposition methods depend on global context embeddings that degrade at extreme scales, and hierarchical methods incur prohibitive runtimes (SIL PRC$_{1000}$: 42h at TSP-100K).

\smallskip\noindent\textbf{Training efficiency.}
\DyNACO{} operates on fixed-size $K$-NN neighborhoods ($K{=}32$), so the GNN processes each neighborhood identically regardless of $N$.
GPU memory is proportional to $n \cdot K$, not $n^2$, and per-ant sampling cost is $O(M \cdot K)$, independent of problem size.
At TSP-100K, DyNACO training completed in roughly 4 hours, while LEHD \cite{lehd} for example required 2.7 days.

\end{document}